\documentclass{article}

\usepackage[preprint]{arxiv}
\usepackage{natbib}
\usepackage[utf8]{inputenc} 
\usepackage[T1]{fontenc}    
\usepackage{hyperref}       
\usepackage{url}            
\usepackage{booktabs}       
\usepackage{amsfonts}       
\usepackage{amsmath}
\usepackage{amssymb}
\usepackage{overpic}

\usepackage{nicefrac}       
\usepackage{microtype}      
\usepackage{xcolor}         
\usepackage{comment}
\usepackage{lipsum}
\usepackage{multirow}
\usepackage{float}
\usepackage{graphicx}
\usepackage{etoolbox}
\usepackage{caption}
\usepackage{svg}
\usepackage{gensymb}
\usepackage{multibib}
\usepackage{wrapfig}
\usepackage{algorithm}
\usepackage{algpseudocode}
\usepackage{xcolor} 
\usepackage{enumitem}
\usepackage{titletoc}
\usepackage{titlesec}

\usepackage{makecell}
\usepackage{array}

\algnewcommand\algorithmicoutput{\textbf{Output:}}
\algnewcommand\Output{\item[\algorithmicoutput]}
\algrenewcommand\textproc[1]{\texttt{#1}}
\newcommand{\method}{ProxyPose}
\newcommand{\projecturl}{https://ruihangzhang97.github.io/proxypose/}
\newcommand{\webpage}{\href{\projecturl}{\color{blue}Project Webpage}}

\newcommand{\cb}{\(\checkmark\)}
\newcommand{\metric}[2]{\makecell[c]{#1$\downarrow$\\[-1pt]{\scriptsize (#2)}}}
\newcommand{\metricup}[2]{#1 $\uparrow$ (#2)}

\newcommand{\vheader}[1]{\makecell[t]{\rotatebox[origin=c]{90}{\strut #1}}}

\providecommand{\Reals}{\mathbb{R}}

\providecommand{\SO}[1]{\mathrm{SO}(#1)}
\providecommand{\identity}{\mathbf{I}}
\providecommand{\Normal}{\mathcal{N}}

\providecommand{\nframes}{N}                  
\providecommand{\nprompts}{Q}                 
\providecommand{\ntokens}{T}                  
\providecommand{\latentdim}{D}                
\providecommand{\dof}[1]{#1\text{-DoF}}       

\providecommand{\src}[1][]{\mathbf{v}\ifx\\#1\\\else_{#1}\fi}
\providecommand{\proxy}[1][]{\mathbf{p}\ifx\\#1\\\else_{#1}\fi}
\providecommand{\proxyhat}[1][]{\hat{\mathbf{p}}\ifx\\#1\\\else_{#1}\fi}
\providecommand{\proxyq}[1]{\proxyhat^{(#1)}}  

\providecommand{\latent}{\mathbf{z}}
\providecommand{\srclatent}{\latent_{\mathrm{src}}}
\providecommand{\proxylatent}{\latent_{\mathrm{proxy}}}
\providecommand{\jointlatent}{\latent_{\mathrm{joint}}}
\providecommand{\noisedlatent}{\tilde{\latent}} 
\providecommand{\firstframetokens}{\mathcal{F}_{1}}

\providecommand{\noise}{\boldsymbol{\epsilon}}
\providecommand{\timestep}{t}
\providecommand{\anchoroffset}{\Delta_\text{offset}} 
\providecommand{\anchortimestep}{\timestep_{\text{offset}}}
\providecommand{\flowalpha}[1][\timestep]{\alpha_{#1}} 
\providecommand{\flowsigma}[1][\timestep]{\sigma_{#1}} 

\providecommand{\generator}{\mathcal{G}_{\theta}}      
\providecommand{\tracker}{\mathcal{T}}                 
\providecommand{\encoder}{\mathcal{E}}                 
\providecommand{\decoder}{\mathcal{D}}                 
\providecommand{\proj}{\pi}                            

\providecommand{\velocitynet}{\boldsymbol{\nu}_{\theta}} 
\providecommand{\velocitytarget}{\boldsymbol{\nu}^{*}} 
\providecommand{\lossweight}[1]{w(#1)} 
\providecommand{\Expected}{\mathbb{E}} 
\providecommand{\loss}{\mathcal{L}}

\providecommand{\rot}{\mathbf{R}}
\providecommand{\trans}{\mathbf{t}}
\providecommand{\rotf}[1][n]{\rot_{#1}}
\providecommand{\transf}[1][n]{\trans_{#1}}
\providecommand{\rotq}[2][n]{\rot_{#1}^{(#2)}} 
\providecommand{\transq}[2][n]{\trans_{#1}^{(#2)}}
\providecommand{\posepriorrot}{\hat{\rot}}
\providecommand{\posepriortrans}{\hat{\trans}}
\providecommand{\intrinsics}{\mathbf{K}}
\providecommand{\focal}{f}
\providecommand{\promptpoint}{\mathbf{q}}              
\providecommand{\promptpointq}[1]{\promptpoint^{(#1)}} 
\providecommand{\cubecorner}[1][k]{\mathbf{X}_{#1}}    
\providecommand{\cubecornersset}{\{\cubecorner\}_{k=1}^{8}}
\providecommand{\imgpoint}{\mathbf{x}}                 
\providecommand{\corrset}[1][n]{\mathcal{C}_{#1}}      
\providecommand{\corrsetq}[2][n]{\mathcal{C}_{#1}^{(#2)}}

\providecommand{\facevertices}[1]{\mathcal{V}_{#1}} 

\providecommand{\depthscalar}[1]{s^{(#1)}}             
\providecommand{\baLoss}{\mathcal{L}_{\mathrm{BA}}}
\providecommand{\wtrans}{w_{\mathrm{t}}}
\providecommand{\wrot}{w_{\mathrm{r}}}
\newcommand{\relrot}[1]{\Delta\mathbf{R}^{(#1)}}
\newcommand{\reltrans}[1]{\Delta\mathbf{t}^{(#1)}}

\usepackage[colorinlistoftodos]{todonotes}

\title{ProxyPose: 6-DoF Pose Tracking\\ via Video-to-Video Translation}

\author{
\makebox[\textwidth][c]{\hspace*{-0.5cm}
Ruihang Zhang$^{*1}$,
Felix Taubner$^{*1,2}$,
Pooja Ravi$^{1}$,
Kiriakos N. Kutulakos$^{1,2}$,
David B. Lindell$^{1,2}$}\\[0.5em]
$^{1}$University of Toronto $^{2}$Vector Institute\\[0.5em]
\small{* Equal contribution}
}

\begin{document}

\maketitle

\setlength{\abovedisplayskip}{4pt}
\setlength{\belowdisplayskip}{4pt}
\setlength{\abovedisplayshortskip}{3pt}
\setlength{\belowdisplayshortskip}{3pt}

\begin{figure}[h!]
\vspace*{-0.3in}
\scriptsize
\centering
\setlength{\tabcolsep}{0pt}
\begin{tabular}{c@{\hspace*{3pt}}c@{\hspace*{3pt}}c}
initial frame \& query pixel 
&
original video frames, overlaid 6-DoF poses \& generated proxy frames 
&
6-DoF pose track
\\	
\includegraphics[width=1.2178in]{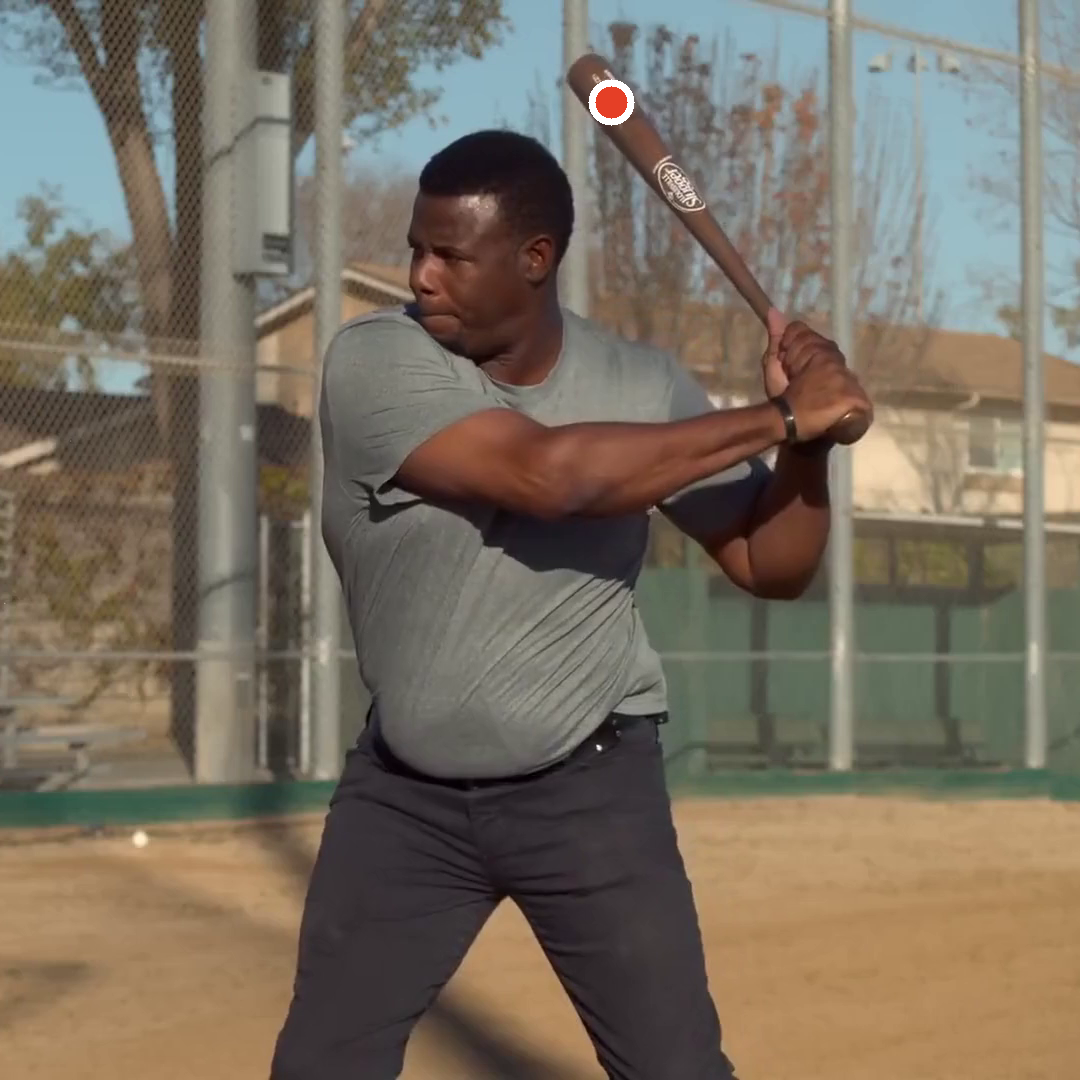} 
&
\raisebox{0.56in}{\begin{tabular}{c@{\hspace*{0.5pt}}c@{\hspace*{0.5pt}}c@{\hspace*{0.5pt}}c@{\hspace*{0.5pt}}c}
\includegraphics[width=0.59in]{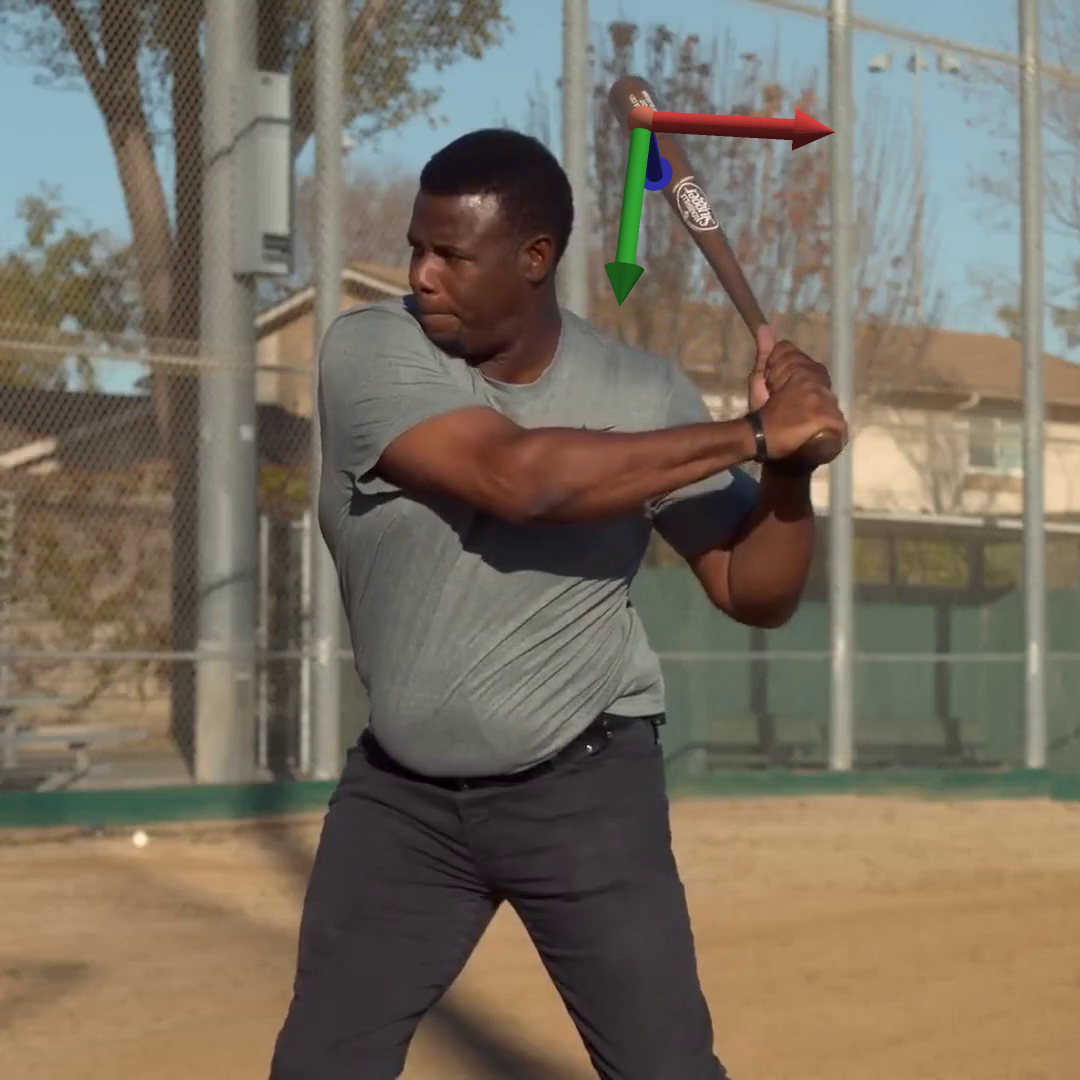}
&
\includegraphics[width=0.59in]{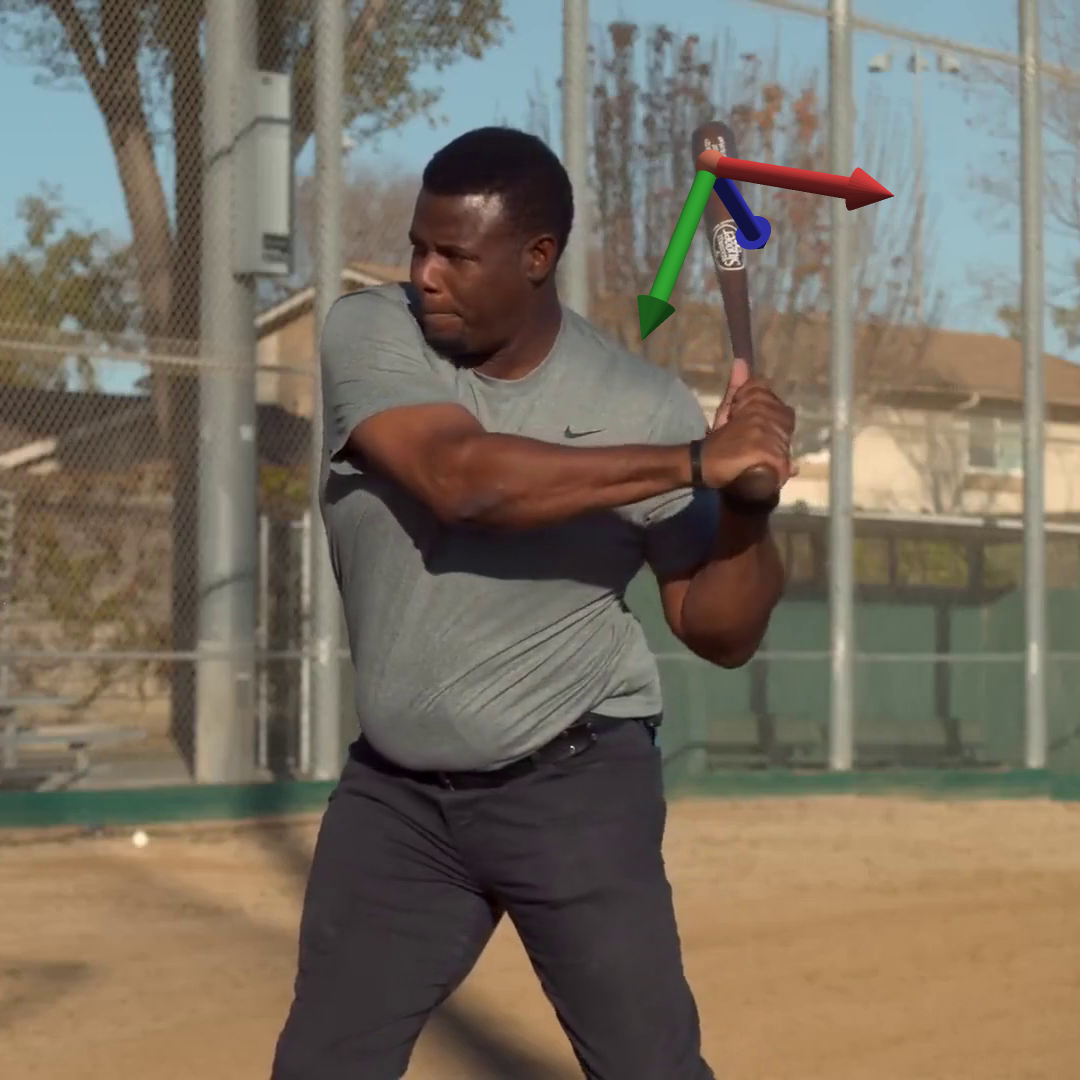}
&
\includegraphics[width=0.59in]{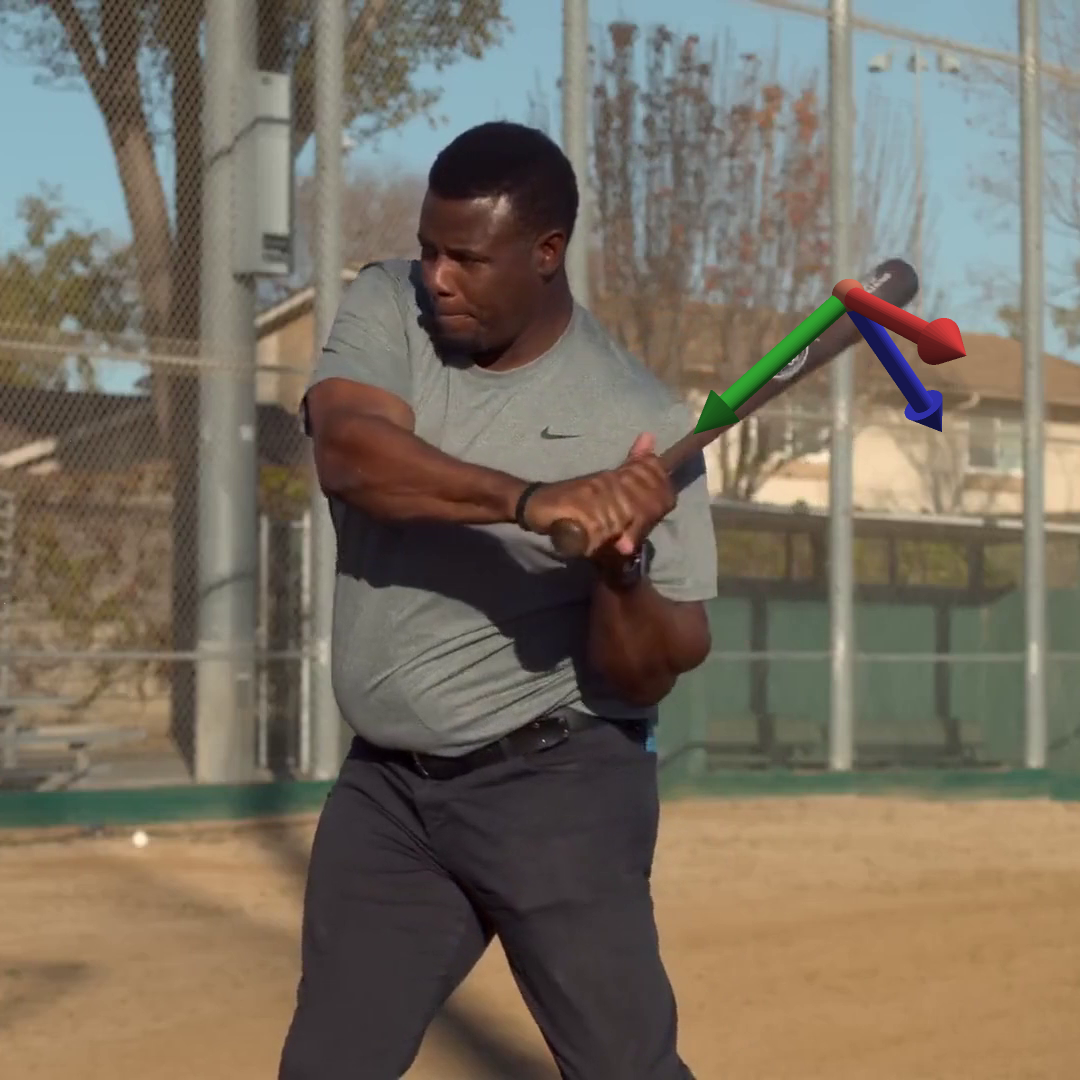}
&
\includegraphics[width=0.59in]{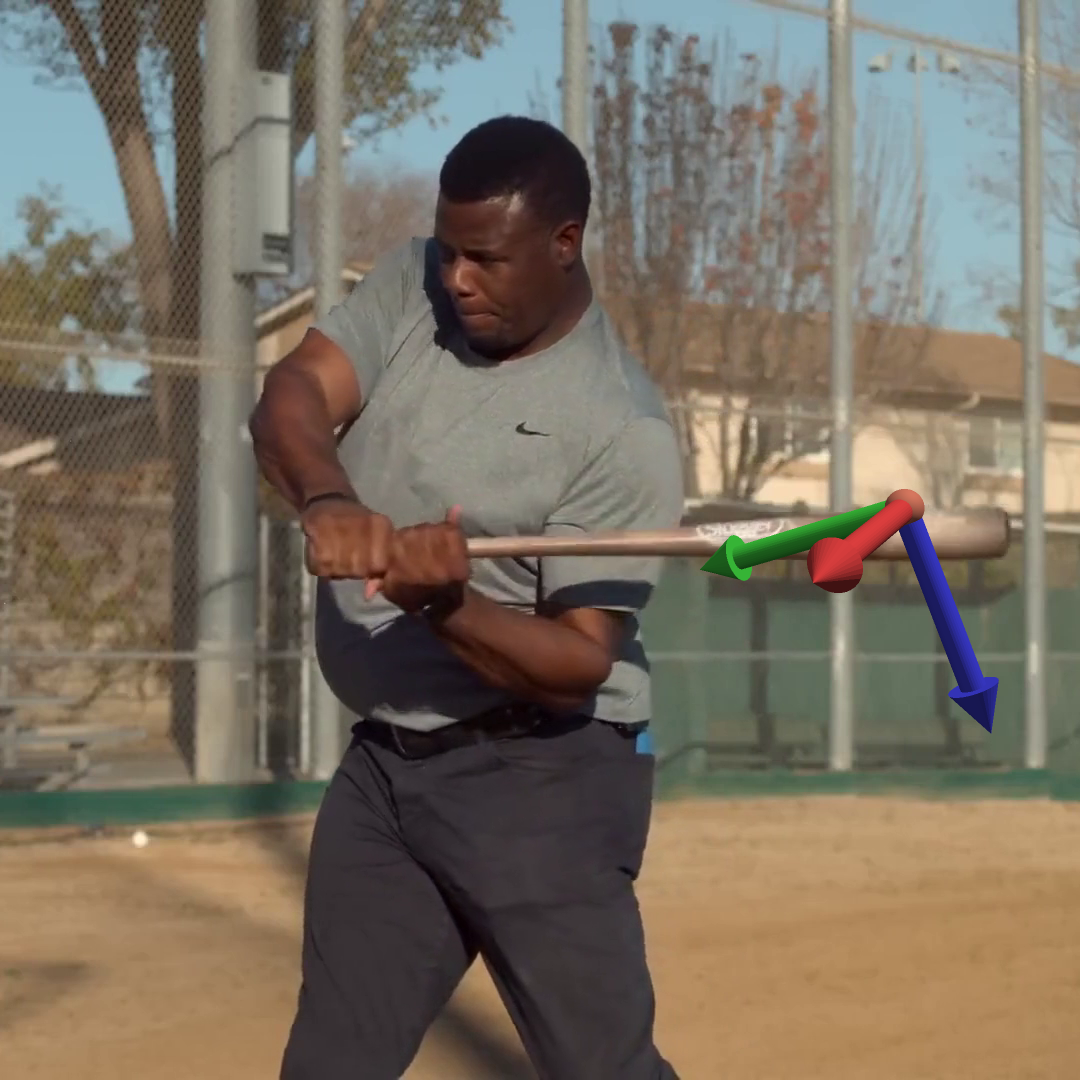}
&
\includegraphics[width=0.59in]{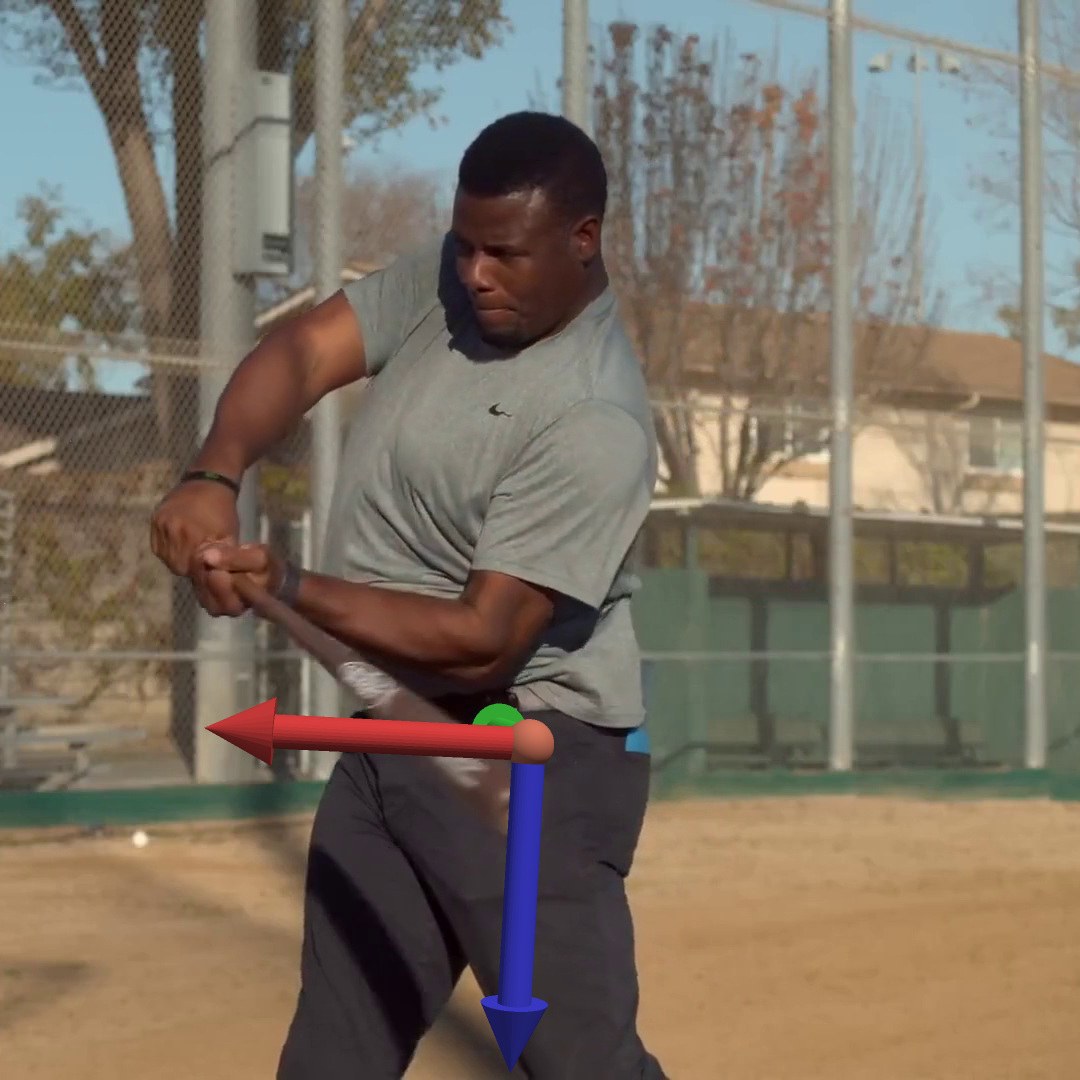}
\\[-0.5pt] 
\includegraphics[width=0.59in]{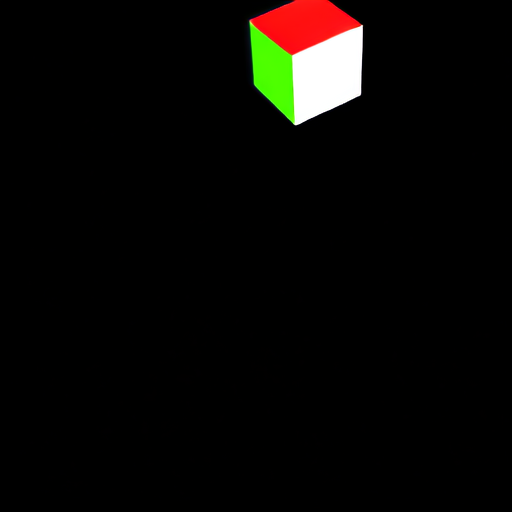}
&
\includegraphics[width=0.59in]{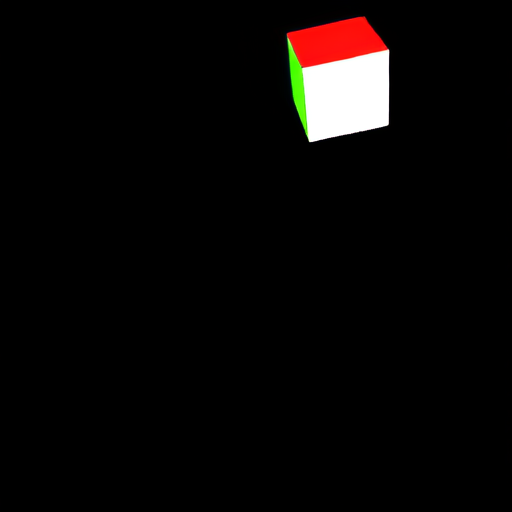}
&
\includegraphics[width=0.59in]{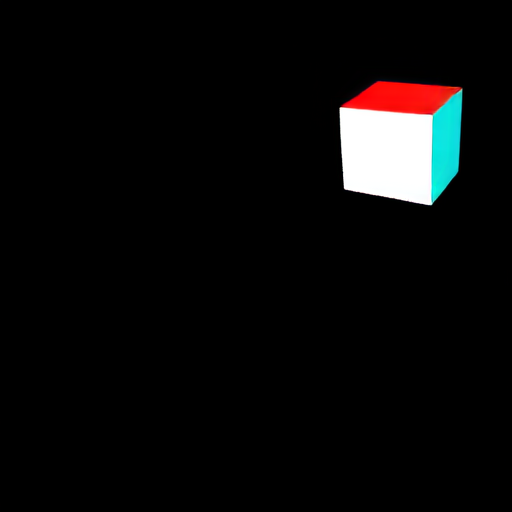}
&
\includegraphics[width=0.59in]{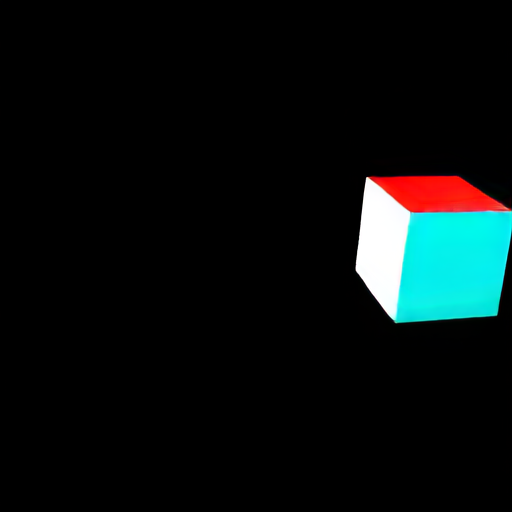}
&
\includegraphics[width=0.59in]{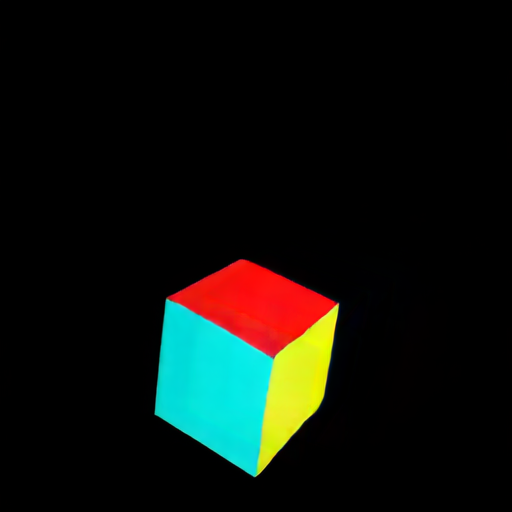}
\end{tabular}}
&
\includegraphics[width=1.2178in]{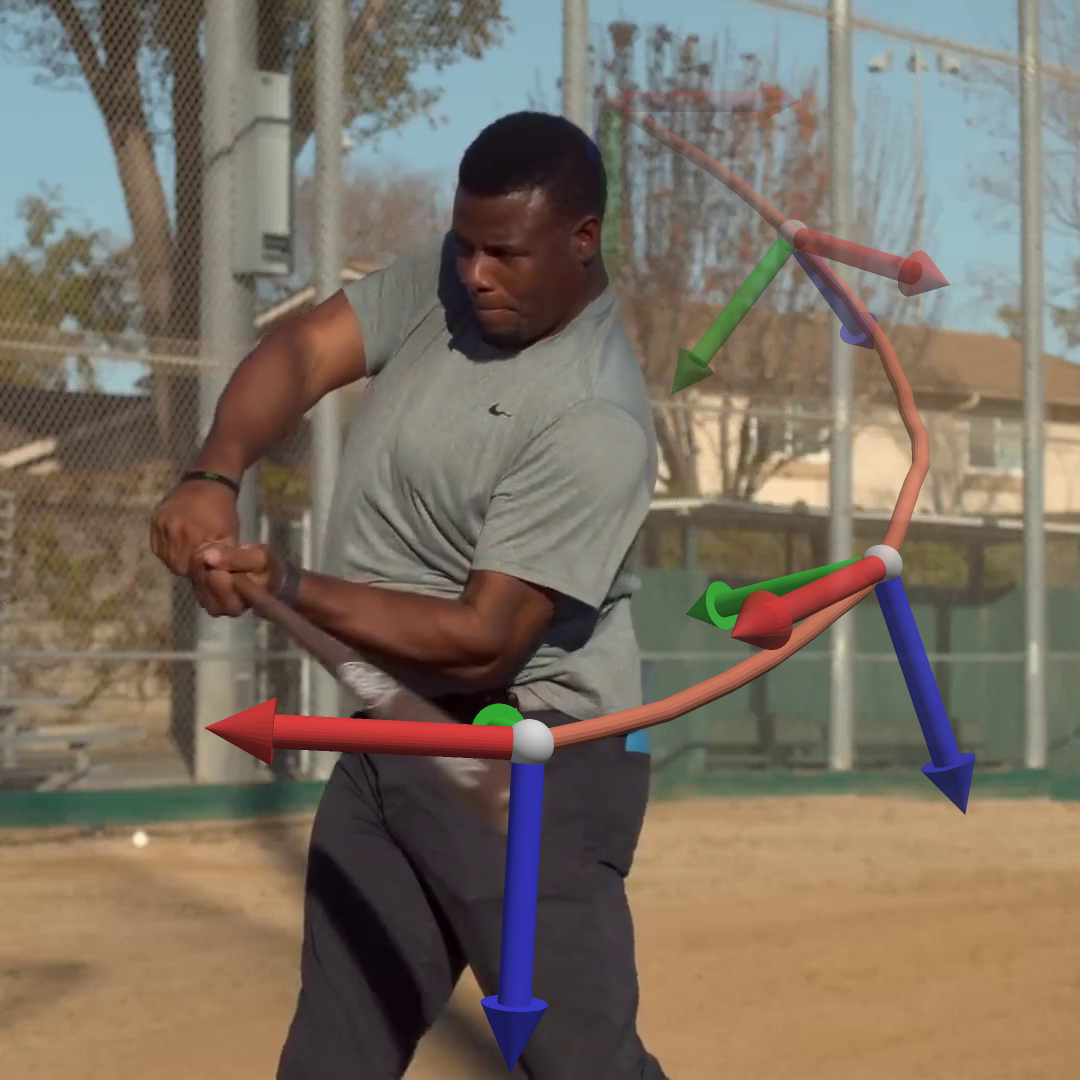} 
\\

\includegraphics[width=1.2178in]{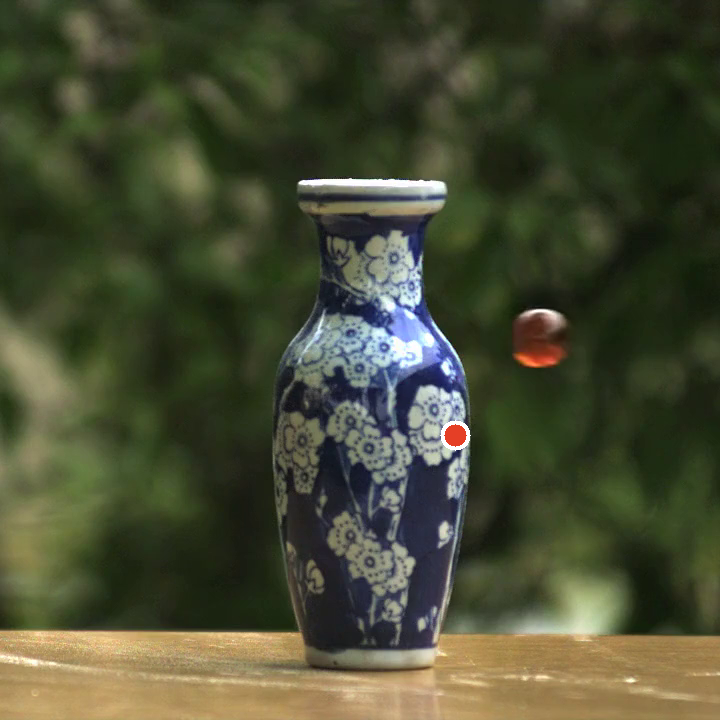} 
&
\raisebox{0.56in}{\begin{tabular}{c@{\hspace*{0.5pt}}c@{\hspace*{0.5pt}}c@{\hspace*{0.5pt}}c@{\hspace*{0.5pt}}c}
\includegraphics[width=0.59in]{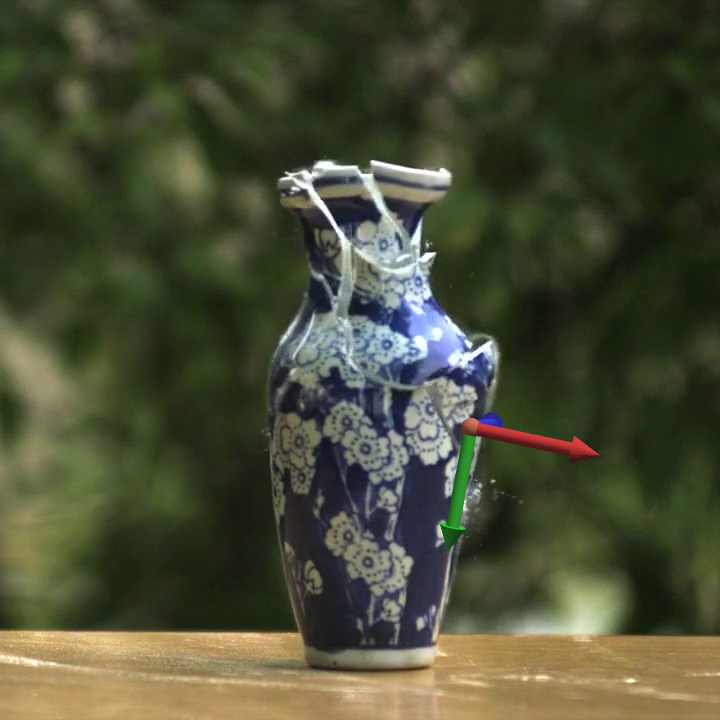}
&
\includegraphics[width=0.59in]{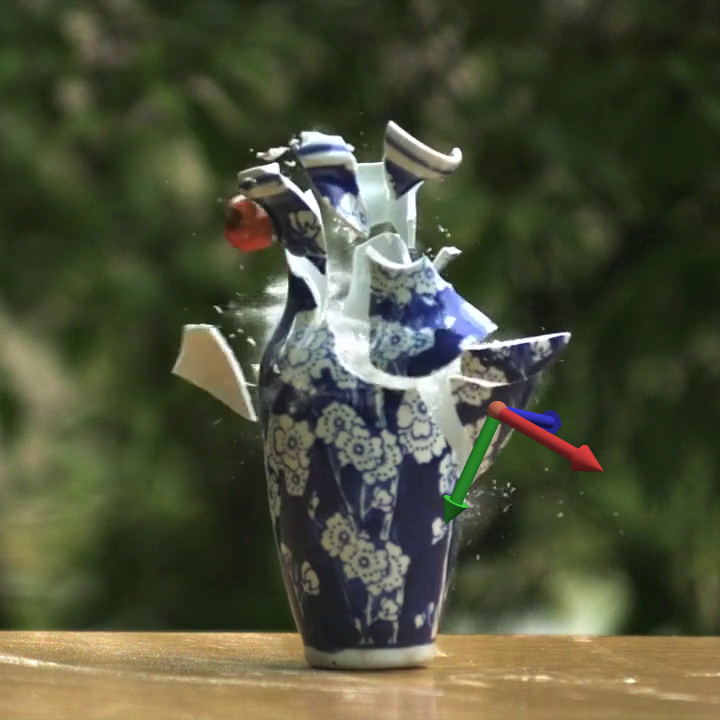}
&
\includegraphics[width=0.59in]{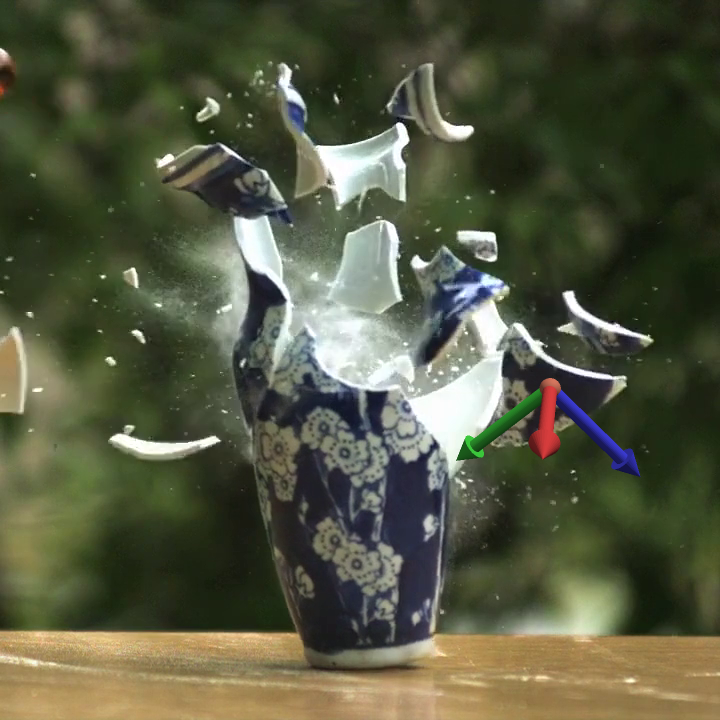}
&
\includegraphics[width=0.59in]{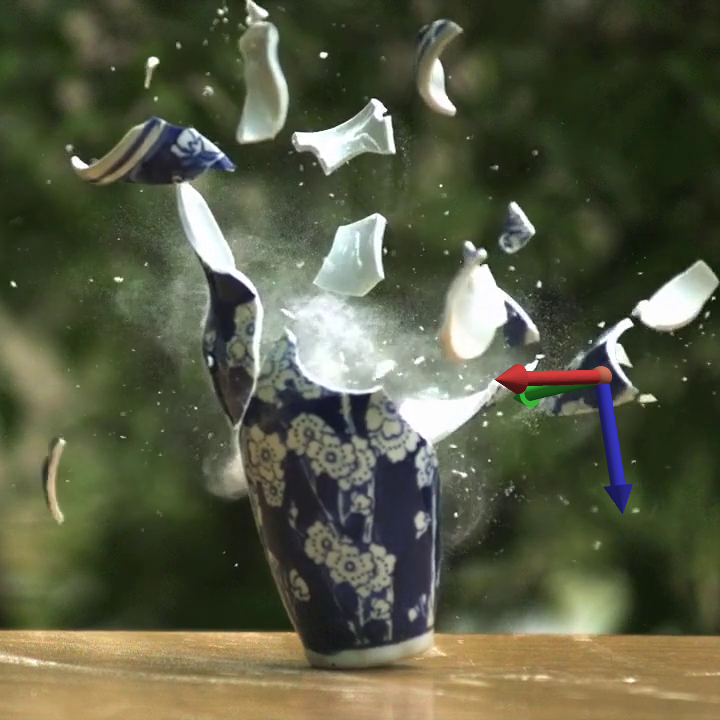}
&
\includegraphics[width=0.59in]{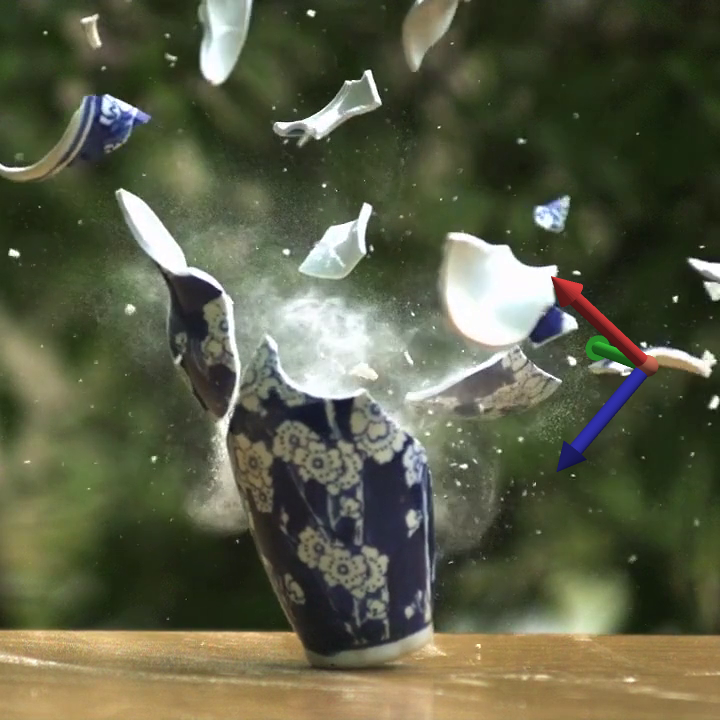}
\\[-0.5pt] 
\includegraphics[width=0.59in]{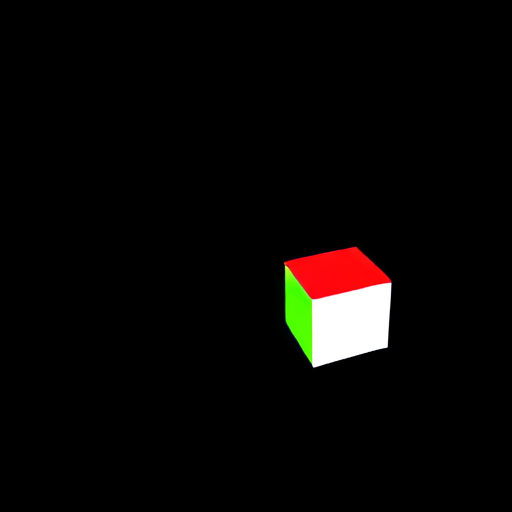}
&
\includegraphics[width=0.59in]{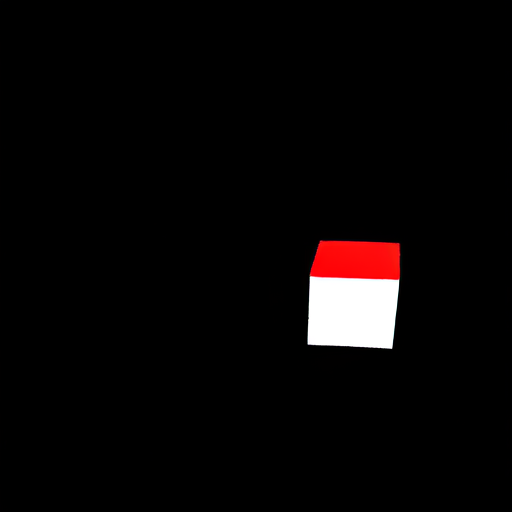}
&
\includegraphics[width=0.59in]{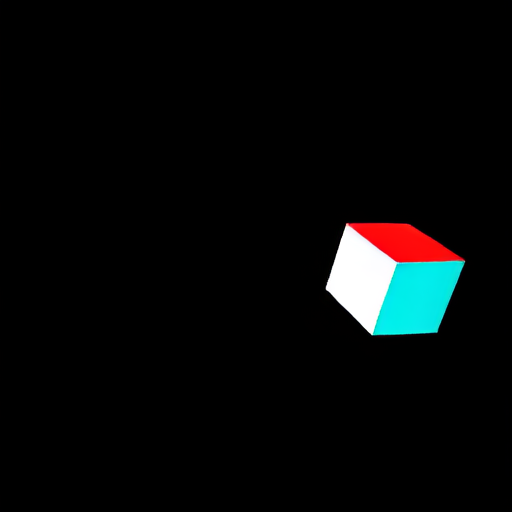}
&
\includegraphics[width=0.59in]{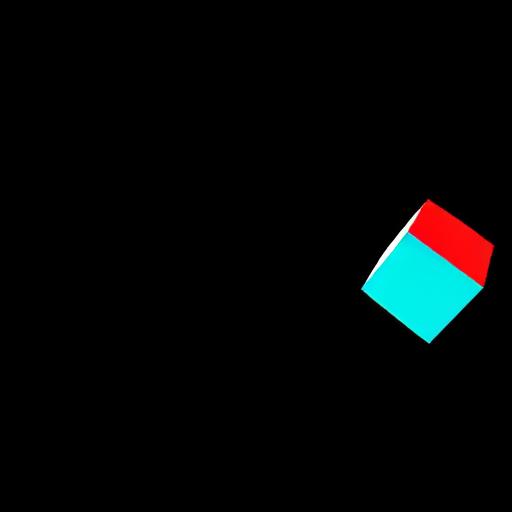}
&
\includegraphics[width=0.59in]{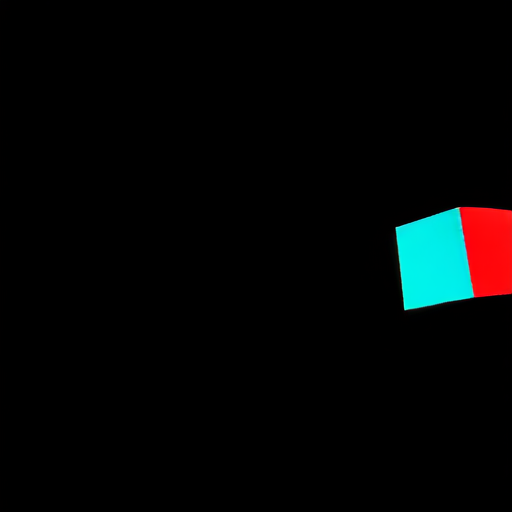}
\end{tabular}}
&
\includegraphics[width=1.2178in]{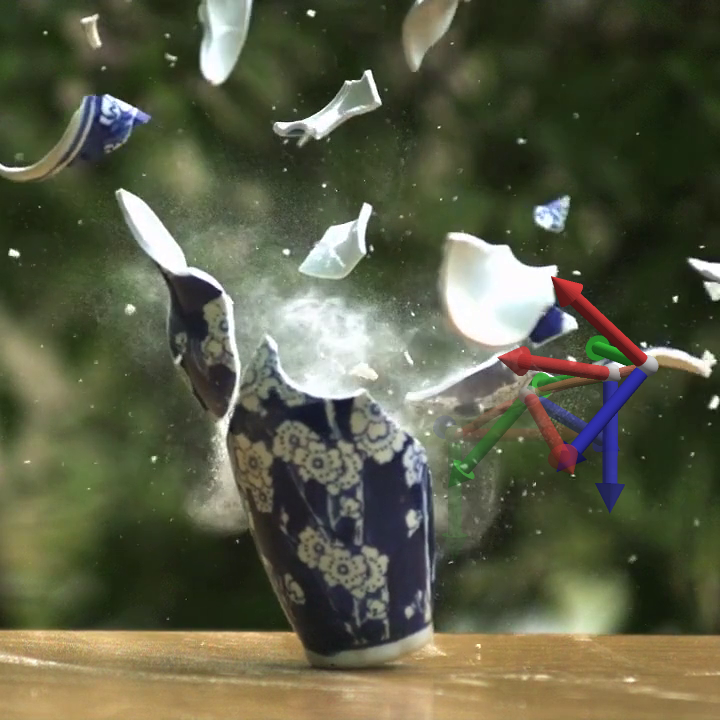} 
\\ 
\includegraphics[width=1.2178in]{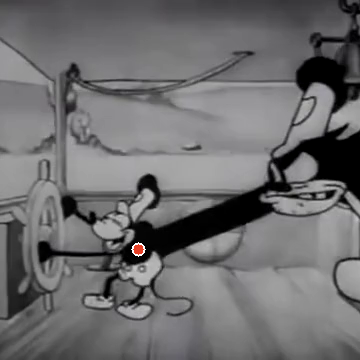} 
&
\raisebox{0.56in}{\begin{tabular}{c@{\hspace*{0.5pt}}c@{\hspace*{0.5pt}}c@{\hspace*{0.5pt}}c@{\hspace*{0.5pt}}c}
\includegraphics[width=0.59in]{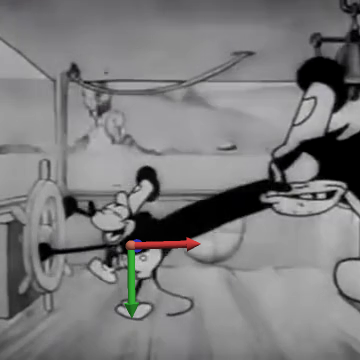}
&
\includegraphics[width=0.59in]{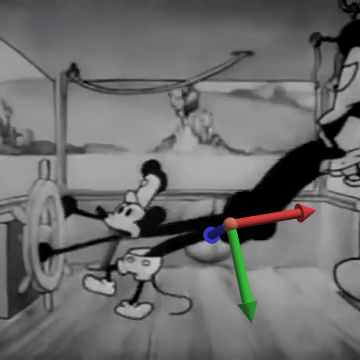}
&
\includegraphics[width=0.59in]{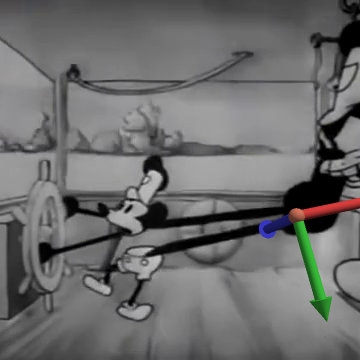}
&
\includegraphics[width=0.59in]{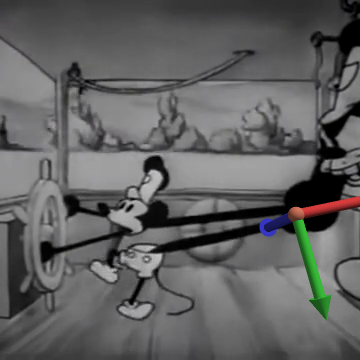}
&
\includegraphics[width=0.59in]{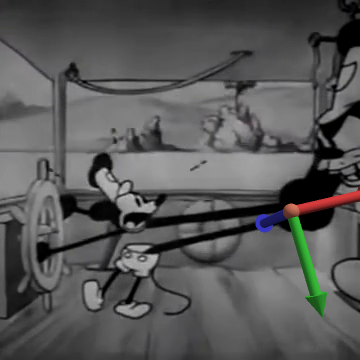}
\\[-0.5pt] 
\includegraphics[width=0.59in]{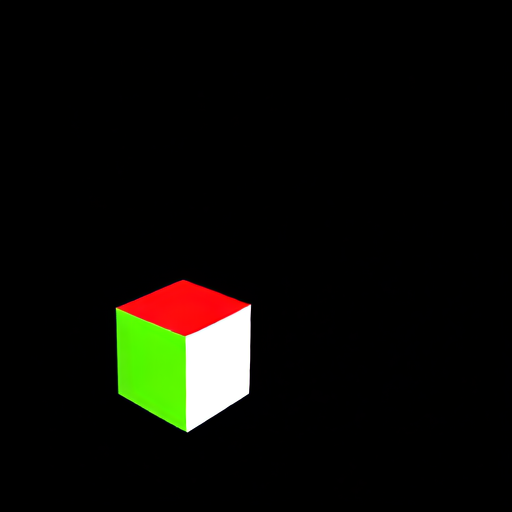}
&
\includegraphics[width=0.59in]{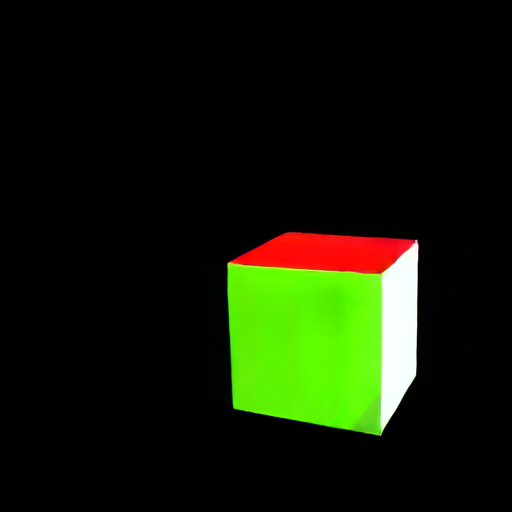}
&
\includegraphics[width=0.59in]{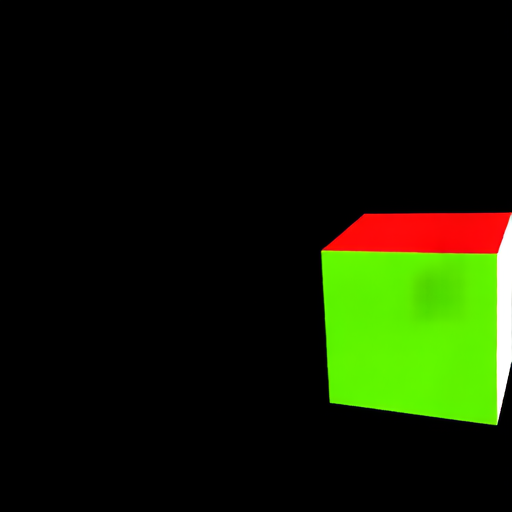}
&
\includegraphics[width=0.59in]{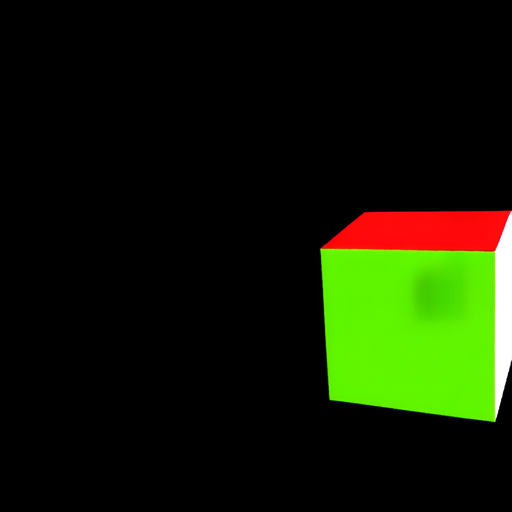}
&
\includegraphics[width=0.59in]{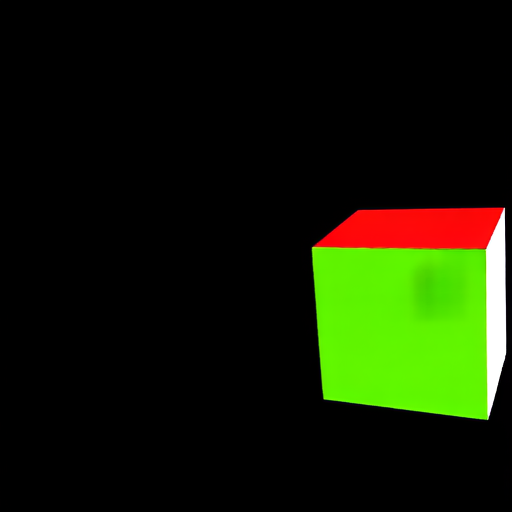}
\end{tabular}}
&
\includegraphics[width=1.2178in]{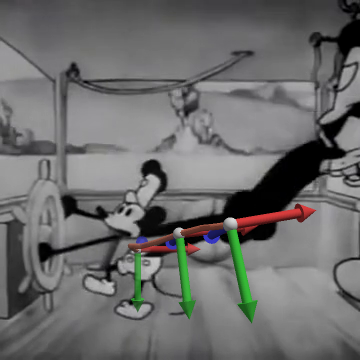} 	
\end{tabular}	
\caption{
Our approach enables tracking relative 6-DoF pose in diverse, highly dynamic scenes without employing foundation models or 3D inference pipelines---including the top of a swinging baseball bat (top); individual regions on a porcelain vase as it fractures (middle); and even Captain Pete's hand in this cartoon clip from \emph{Steamboat Willie} (bottom). Highlighted points are the proxy cube's center. We use this core capability to track multiple surface regions as they move, deform or occlude each other in unconstrained internet videos (\textbf{see the videos on the \webpage{} for several such examples}).}
\label{fig:teaser}
\end{figure}
\begin{abstract}
Tracking the six-degree-of-freedom (6-DoF) pose of objects and surfaces from monocular video is a long-standing problem in computer vision. 
To tackle this problem, existing methods require inputs beyond the video itself---such as 3D models, depth maps, object masks, or task-specific learned features---and they struggle with textureless, transparent, reflective, or deformable surfaces.
Here, we introduce \method{}, which recasts 6-DoF pose tracking as video-to-video translation.
Given only a video and a single marked pixel in the first frame, a fine-tuned video diffusion model translates the input into a \emph{proxy video}---a synthetic video depicting a colored polyhedron undergoing the same local rigid-body motion as the surface region at the marked pixel.
Because the proxy's geometry and appearance are known by construction, recovering its full 6-DoF trajectory reduces to classical pose estimation with off-the-shelf solvers.
This formulation leverages large-scale video pre-training to absorb the hardest aspects of pose tracking---handling challenging materials, occlusions, and deformations---into the translation step, while operating at the pixel level with no assumptions about object identity, boundaries, or global rigidity.
\method{} achieves state-of-the-art 6-DoF pose tracking accuracy without the additional inputs required by competing methods and after fine-tuning the video model only on synthetic data.
We further demonstrate that \method{} extends to face tracking, camera pose estimation, and challenging in-the-wild scenes that are beyond the reach of existing approaches. Video results are available on the \webpage{}.

\end{abstract}

\vspace{-1em}
\section{Introduction}
\label{sec:introduction}

Tracking the six-degree-of-freedom (6-DoF) pose of objects and surfaces from video is a fundamental problem in computer vision that traces its roots back more than thirty years.
Classical techniques spanned a range of model-based approaches, from methods that aligned rigid CAD models to images~\citep{lowe1992fitting,harris1993tracking,drummond2002realtime,comport2006accurate} to deformable-template and statistical-shape approaches developed primarily for non-rigid face and human-body tracking~\citep{cootes1998active,bregler1998tracking,blanz1999morphable,decarlo2000optical}, to 
tracking-by-detection methods
which leveraged discriminative classifiers or convolutional neural networks~\citep{shotton2010kinect,hinterstoisser2012model,kehl2017ssd6d,xiang2018posecnn}.
A complementary line of research combined 2D feature detection~\citep{lucas1981iterative,shi1994good,lowe2004distinctive} with structure from motion (SfM) to simultaneously recover 3D models and pose of rigid~\citep{fitzgibbon1998sfm} or non-rigid~\citep{torresani2001nonrigidsfm} scenes by global bundle adjustment~\citep{triggs1999bundle,agarwal2009rome} or sequential state estimation~\citep{davison2007monoslam,engel2014lsd,murORB2}. This general approach has been extended to include neural-network predictions for improved 6-DoF tracking flexibility, robustness and efficiency~\citep{he2022onepose,sun2022onepose}. All these approaches, however, still rely on explicit 3D representations, reference images, and task-specific features.

Most recently, 
the development of foundation models for object segmentation~\citep{kirillov2023segment,ravi2024sam2}, monocular depth inference~\citep{yang2024depth,yang2024depthv2,ke2024repurposing}, 
pose estimation~\citep{wen2024foundationpose},
dense point tracking~\citep{karaev2024cotracker,doersch2023tapir,zhangtapip3d}, and feed-forward scene reconstruction~\citep{wang2025vggt,leroy2024mast3r,li2025megasam} is opening the door to pipelines that build on these models to achieve even more robust 3D tracking, reconstruction, and camera-pose estimation from general videos.
Such models typically involve large, specially curated datasets and advanced training curricula to achieve robustness on generic mid-level vision tasks such as segmentation, matching, and depth prediction.

In this work, we step back to ask a broader question: are such large-scale, task-specific foundation models \emph{necessary} for robust 3D object tracking?
We provide preliminary evidence that suggests the answer may be \emph{no}.
Specifically, we show that large video generation models~\citep{ho2022video,blattmann2023stable,videoworldsimulators2024,polyak2025moviegen}---already trained on billions of images and millions of videos---provide an alternative route to 3D object tracking and pose estimation, without any explicit object segmentation, feature tracking, or 3D scene reconstruction.

Our approach recasts 6-DoF tracking as a \emph{video-to-video translation} problem that can be tackled by fine-tuning a large pre-trained video diffusion model.
Given an input video and a single marked pixel in the first frame, our model translates the input into a \emph{proxy video}: a synthetic video depicting a simple, known CAD primitive (e.g., a colored polyhedron) undergoing the same 3D motion as the surface region at the marked pixel, and rendered against a black background.
Because the primitive's appearance is designed for easy detection, recovering 6-DoF motion reduces to CAD-based pose estimation
on the proxy video---a problem that is easily handled with classical algorithms.

The key insight underlying our approach is that large video models already \emph{implicitly} encode rich information about object surfaces and their movement in three dimensions---including how rigid and non-rigid 3D motion manifests as 2D appearance change. This formulation offers several concrete advantages over existing pipelines:
(1)~it operates at the \emph{pixel level}, making no assumptions about object identity, boundaries, or rigidity, and thereby sidestepping potentially unreliable or training-heavy segmentation stages;
(2)~it requires no 3D models, depth sensors, precomputed feature representations, or large-scale curated datasets;
(3)~it handles non-rigid, textureless, or shiny surfaces without any special-case treatment;
and (4)~it inherits the temporal consistency of the underlying video model, enabling tracking through occlusions and rapid motion.

We evaluate our approach, \textbf{\method}, on a range of challenging scenarios and demonstrate that it achieves competitive or superior accuracy and temporal consistency compared with state-of-the-art baselines---without the explicit 3D representations, segmentations, or task-specific training those methods rely on.
 Beyond local region tracking, we show that the recovered per-pixel \dof{6} trajectories can be aggregated to obtain camera-pose estimates and to perform face tracking, illustrating the broader utility of pixel-level 6-DoF motion representations.

\vspace{-0.5em}
\paragraph{Overview of limitations.} Since our approach relies on an off-the-shelf video model, the number of frames that can be processed in a single pass is bounded by the generation length of the model, and only \emph{relative} motion can be recovered---absolute pose requires introducing additional constraints.
Nonetheless, these findings point toward a promising and largely unexplored direction in which generative video models serve as a general-purpose backbone for 3D motion understanding.

\vspace{-1em}
\section{Related Work}

\vspace{-0.5em}
\paragraph{6-DoF object pose estimation and tracking.}

The literature on 6-DoF pose estimation spans classical geometric solvers~\citep{lowe2004distinctive,lepetit2009epnp,besl1992icp,hinterstoisser2012model}, instance- and category-level deep networks~\citep{kehl2017ssd6d,xiang2018posecnn,peng2019pvnet,wang2019densefusion,wang2019nocs,labbe2023megapose,zhang2024omni6dpose}, and, more recently, \emph{model-free} methods that remove the requirement for explicit 3D models during training and inference.
Among the latter, Gen6D~\citep{liu2022gen6d}, OnePose~\citep{sun2022onepose}, and FoundationPose~\citep{wen2024foundationpose} estimate pose from reference images or local reconstructions, while subsequent work introduces single-view matching~\citep{he2024nope,corsetti2024oryon}, vision-language-model-based reasoning~\citep{kuang2026conceptpose}, and image-to-3D generation pipelines for downstream pose recovery~\citep{nguyen2024gigapose,liu2025hippo,pan2025omnimanip}.
Although previous model-free approaches relax the requirement for explicit 3D models, they still require task-specific training regimens and depend on extracting precisely localizable features or markings from the observed object, and hence are sensitive to textureless, transparent, or deformable surfaces.

\vspace{-0.5em}
\paragraph{Point tracking.}
Tracking arbitrary points across video frames is a long-established problem, originating in sparse feature tracking~\citep{lucas1981iterative,shi1994good} and optical-flow estimation~\citep{horn1981determining}.
The recent \emph{tracking any point} (TAP) formulation~\citep{doersch2022tapvid} has spurred a new generation of dense point trackers.
Foundation models such as TAPIR~\citep{doersch2023tapir} and CoTracker~\citep{karaev2024cotracker} leverage large-scale training and transformer architectures to achieve accurate 2D correspondences even through occlusions.
SpatialTracker~\citep{xiao2024spatialtracker} and TAPIP3D~\citep{zhangtapip3d} extend this paradigm to 3D by lifting tracked points using monocular depth estimators. Our method complements point tracking by recovering per-pixel 6-DoF trajectories from a single marked pixel.

\vspace{-0.5em}
\paragraph{Controllable video generation.}
Image and video generation have advanced rapidly with diffusion-based architectures~\citep{rombach2022high,ho2022video,blattmann2023stable,videoworldsimulators2024,polyak2025moviegen}.
A growing body of work controls these generative models through auxiliary conditioning signals---edge maps, depth, segmentation, and human pose---using architectures such as ControlNet~\citep{zhang2023controlnet} and related designs~\citep{wang2024motionctrl,taubner2025cap4d,taubner2025mvp4d,bahmani2026lyra}.
Conversely, recent methods invert the generative process for discriminative tasks: repurposing diffusion models as structured priors for monocular depth estimation~\citep{ke2024repurposing,fu2024geowizard}, surface-normal prediction~\citep{ye2024stablenormal}, optical-flow estimation~\citep{saxena2023surprising}, and neural inverse rendering~\citep{liang2025diffusionrenderer}.
\citet{tedla2025blur2vid} demonstrate that a video diffusion model can reconstruct sharp video from a single motion-blurred image, further illustrating that these models encode rich motion priors.

\vspace{-0.5em}
\paragraph{Video-to-video translation.}
Diffusion models have been widely adopted for translating an input video into a modified output for applications in editing~\citep{mai2026easyv2v,zhang2025v2edit,liang2025streamv2v,jiang2025vace}, style transfer~\citep{ye2025stylemaster}, control~\citep{geng2025motionprompting}, and novel-view synthesis~\citep{jeong2025reangle}. 
Most closely related to our work, Point Prompting~\citep{shrivastava2025pointprompting} shows that pre-trained video diffusion models can perform zero-shot 2D point tracking by placing a colored marker at a query point and regenerating the video with the tracked marker.
Our approach shares the insight that video diffusion models encode strong motion priors, but rather than recovering 2D point trajectories, we fine-tune the model to produce structured \emph{proxy videos} that reveal rich 3D information about 6-DoF pose and motion.
To our knowledge, recovering 6-DoF pose from video through video-to-video translation with diffusion models has not been previously explored.
 
\vspace{-0.5em}
\section{Method}
\label{sec:method}
\vspace{-0.5em}

Given a source video $\src = \{\src[n]\}_{n=1}^{\nframes}$ of $\nframes$ frames and a single marked query pixel $\promptpoint \in \Reals^{2}$ in the first frame, we recover per-frame rotations $\rotf \in \SO{3}$ and translations $\transf \in \Reals^{3}$ via
\begin{equation}
    \proxyhat \;=\; \generator(\src,\, \promptpoint)
    \qquad \text{and} \qquad
    \{(\rotf,\, \transf)\}_{n=1}^{\nframes}
    \;=\; \tracker(\proxyhat)\ .
    \label{eq:overview}
\end{equation}

\begin{figure}[t]
    \centering
    \includegraphics[width=\textwidth]{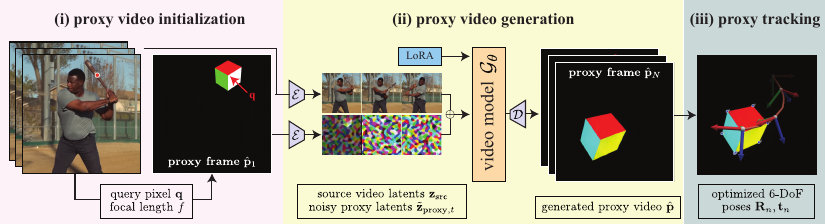}
    \vspace{-1.5em}
    \caption{\method{} pipeline overview. Given a source video and a single marked query pixel on a target surface region, we first translate the input into a \emph{proxy video} in which a colored polyhedron undergoes the same local rigid-body motion (i-ii), and then recover the polyhedron's \dof{6} pose trajectory (iii) by Perspective-$n$-Point (PnP) and, optionally, multi-query bundle adjustment.}
    \label{fig:pipeline}
    \vspace{-1.5em}
\end{figure}

$\generator$ in Equation~\ref{eq:overview} 
is a fine-tuned video diffusion model that synthesizes a proxy video $\proxyhat = \{\proxyhat_n\}_{n=1}^{\nframes}$, \mbox{and $\tracker$ is} a deterministic tracker that exploits the known geometry and appearance of the proxy object (see Figure~\ref{fig:pipeline}). We use
a cube with faces of different color placed on a black background in all experiments.We assume that the camera's focal length is known (or can be coarsely approximated), and that the sensor has a known aspect ratio and square pixels. The camera intrinsic matrix can therefore be written as $\texttt{diag}(\focal, \focal, 1)$.
Below, we describe the video-to-video translation stage that produces the proxy, and then consider the geometric solver that extracts \dof{6} pose from the proxy.

\vspace{-0.5em}
\subsection{Proxy Video Generation via Video-to-Video Translation}
\label{sec:proxy-generation}

\vspace{-0.5em}
\paragraph{Initializing the proxy video.}
To construct the first frame $\proxyhat_1$
of the proxy video, we render the cube 
so that three of its faces project to 
equal-sized regions in the image, its projection occupies a fixed fraction of the image area, and its non-silhouette vertices project to the query pixel (Figure~\ref{fig:pipeline}(i)). More specifically, we set the cube's initial pose ($\rot_1$, $\trans_1$)
and size as follows.
We first choose the
initial rotation $\rot_1$ so that the cube's $[1,1,1]^\top$ diagonal is aligned with the camera ray through the query pixel $\promptpoint$. We then take the cube's center to be at unit depth along that ray, setting $\trans_1$ accordingly.
To balance the cube's projected area against the image area available for tracking, we set the length of the cube's edges to $(h / \focal)s_\text{cube}$, where $h$ is the image height in pixels, $\focal$ is the focal length and $s_\text{cube}=0.15$. This ensures the proxy occupies a consistent fraction of the image regardless of $h$ or $\focal$. At training time, we render the ground-truth proxy with the known relative object poses $(\rotf, \transf)_{n=1}^{\nframes}$.

\vspace{-0.5em}
\paragraph{Architecture and conditioning.}
We instantiate $\generator$ by fine-tuning a pretrained video diffusion model~\citep{wan2025} with low-rank adaptation (LoRA)~\citep{hu2022lora} on the query, key, value, and output projection matrices of the self- and cross-attention layers as well as the feedforward layers.
The source and proxy video are each encoded by the pretrained variational autoencoder (VAE), yielding latents $\srclatent = \encoder(\src)$ and 
$\proxylatent = \encoder(\proxyhat_1)$. The latents are patchified into spatio-temporal token sequences of length $\ntokens$ and concatenated along the token dimension:
\begin{equation}
    \jointlatent \;=\; [\,\srclatent;\, \proxylatent\,]
    \;\in\; \Reals^{2\ntokens \times \latentdim}.
    \label{eq:joint-latent}
\end{equation}

Joint self-attention over $\jointlatent$ enables the diffusion transformer (DiT) to model long-range correspondences between the observed object motion in $\srclatent$ and the proxy geometry in $\proxylatent$.
To allow the model to distinguish the two streams, we modify the rotary positional embedding from 3D to 4D, where we set the last axis to encode a stream identifier ($-1$ for proxy tokens, $1$ for source tokens). A fixed text prompt describing the desired output is supplied to the cross-attention layers throughout training and inference (see Supp.\ Section~\ref{supp:generation} for additional details).

\vspace{-0.5em}
\paragraph{Noise scheduling.}
Naively adding noise to the entire proxy stream can destabilize the identity, scale, and orientation of the generated proxy, particularly early in sampling.
To mitigate this, we introduce a \emph{noise schedule offset} for the first frame.
Let $\firstframetokens$ denote the set of proxy tokens corresponding to the first video frame.
Rather than corrupting these tokens at the global timestep $\timestep$, we add noise corresponding to a reduced timestep $\anchortimestep = \max(\timestep - \anchoroffset,\, 0)$, where $\anchoroffset$ is a fixed offset:
\begin{equation}
    \noisedlatent_{\text{proxy}, \timestep}^{(i)}
    \;=\;
    \begin{cases}
        \flowalpha[\anchortimestep]\,\proxylatent^{(i)}
        + \flowsigma[\anchortimestep]\,\noise^{(i)},
        & i \in \firstframetokens,\\[4pt]
        \flowalpha[\timestep]\,\proxylatent^{(i)}
        + \flowsigma[\timestep]\,\noise^{(i)},
        & i \notin \firstframetokens,
    \end{cases}
    \label{eq:anchor-schedule}
\end{equation}
where $\flowalpha$ and $\flowsigma$ are the flow-matching signal and noise coefficients, and $\noise^{\,i} \sim \Normal(0, \identity)$ is independent Gaussian noise.
This ensures that the first proxy frame remains close to the data manifold throughout training while still participating in the denoising process, which we find stabilizes video generation.

\vspace{-0.5em}
\paragraph{Fine-tuning.}
We fine-tune $\generator$ with a flow-matching objective~\citep{lipman2023flow,liu2023rectified} in which only the proxy stream is corrupted; the source stream $\srclatent$ remains clean and serves as the motion-conditioning signal.
Training minimizes the standard flow-matching loss over all proxy tokens, with $\srclatent$ concatenated as conditioning:
\begin{equation}
    \loss
    \;=\;
    \Expected_{\timestep,\,\noise}
    \Big[\,
        \lossweight{\timestep}\,
        \big\|\,
            \velocitynet\!\big(\noisedlatent_{\text{proxy},\timestep},\, \srclatent,\, \timestep\big)
            - \velocitytarget
        \big\|_2^2
    \,\Big],
    \label{eq:flow-loss}
\end{equation}
where $\velocitynet$ is the DiT velocity prediction, $\velocitytarget$ is the flow-matching target, and $\lossweight{\timestep}$ is a timestep-dependent loss weight~\citep{lipman2023flow}. At inference, the denoised proxy latents are decoded by the pretrained VAE decoder $\decoder$ to obtain $\proxyhat$, which is passed to the geometric tracking stage.

\vspace{-0.5em}
\subsection{Proxy Video Tracking}
\label{sec:proxy-tracking}
Each of the cube's six faces has a distinct color, and its 3D vertex positions $\cubecornersset$ are known in the cube's local coordinate frame. 
Hence, recovering the per-frame pose $(\rotf,\, \transf)$ given the focal length $\focal$ reduces to detecting quadrilaterals corresponding to each visible face, establishing 2D--3D corner correspondences $(\imgpoint, \cubecorner[])\in \corrset$, and solving PnP~\citep{Li2012rpnp,Zheng2014pnp}.
We provide a high-level summary of the full procedure in Supp.\ Algorithm~\ref{supp:alg:tracking} and pseudocode in Supp.\ Algorithm~\ref{supp:alg:tracking-detailed} (see Supp.\ Section~\ref{supp:tracking}); all image-processing primitives---cube face segmentation, vertex localization, and PnP---are provided by OpenCV~\citep{bradski2000opencv}.
The appearance and geometry of the proxy are straightforward, so this pipeline is reliable despite the absence of any learned components.

\vspace{-0.5em}
\paragraph{Enforcing temporal smoothness.}
After recovering initial per-frame poses using PnP, we refine the pose sequence $\{(\rotf,\, \transf)\}_{n=2}^N$ by minimizing reprojection error and a temporal smoothness penalty:
{\small
\begin{equation}
    \loss_{\mathrm{smooth}} \;=\;
    \sum_{n=1}^{\nframes}
        \!\sum_{(\imgpoint,\, \cubecorner[]) \in \corrset[n]}\!
        \big\lVert \proj(\rotf[n]\,\cubecorner[] + \transf[n]) - \imgpoint \big\rVert_{2}^{2}
    \;+\;
    \sum_{n=1}^{\nframes-1} \!\Big[
        \wtrans \big\lVert \trans_{n+1} \!-\! \transf[n] \big\rVert_{2}^{2}
        + \wrot \big\lVert \log\!\big( \rot_{n+1} \rotf[n]^{\!\top} \big) \big\rVert_{\text{F}}^{2}
    \Big],
\label{eq:smoothness}
\end{equation}
}
\!where $\proj(\mathbf{u}) = (\focal/ u_3) \, [u_1, u_2]^\top$ denotes perspective projection.
The first term penalizes reprojection error across all frames; the second encourages temporal smoothness by penalizing frame-to-frame changes in translation (weighted by $\wtrans$) and rotation (weighted by $\wrot$ and measured via the Frobenius norm of the rotation logarithm).
We optimize the pose sequence with Levenberg--Marquardt, initializing from the per-frame PnP solutions while holding $(\rot_1,\, \trans_1)$ fixed.

\vspace{-0.5em}
\subsection{Incorporating Rigidity Constraints}
\label{sec:rigidity}
\vspace{-0.5em}

For rigid surfaces, multiple query pixels $\{\promptpointq{q}\}_{q=1}^{\nprompts}$ may optionally be placed on them, yielding multiple proxy videos $\{\proxyq{q}\}_{q=1}^{\nprompts}$
whose individual pose estimates can be fused via bundle adjustment to reduce sensitivity to query-specific tracking errors.
As we show in the evaluation, this improves performance for surfaces that are known to be rigid. Such a procedure, however, is not necessary for general in-the-wild tracking with our method, where scene properties are not known a priori.

\paragraph{Multi-query bundle adjustment.}
We run the method of Section~\ref{sec:proxy-tracking} independently on each proxy video $\proxyq{q}$ to obtain per-frame 2D--3D correspondences $\corrsetq[n]{q}$ and an initial pose sequence
for each video. Since every proxy cube is initialized at unit depth (Section~\ref{sec:proxy-generation}), the recovered translations do not reflect the true relative depths of the proxy cubes. To account for their unknown relative depth, we introduce a per-proxy depth scalar $\depthscalar{q}>0$
that must be estimated as part of the bundle adjustment procedure. This scalar essentially controls the perspective depth/scale ambiguity,
redefining the depth of proxy cube $q$ in frame 1 to be
$\depthscalar{q}$ and resizing the cube by $\depthscalar{q}$ so that its perspective projection in frame 1 is unaffected. By convention, we set the scale of the first proxy cube to $\depthscalar{1}=1$.

We parameterize the shared rigid motion of all proxies through the poses $\{(\rotq[n]{1}, \transq[n]{1})\}_{n=1}^{\nframes}$ of the first proxy. 
Specifically, the pose of proxy~$q$ is a fixed rigid transformation of the pose of proxy 1:
\begin{equation}
    \rotq[n]{q} = \rotq[n]{1}\,\relrot{q}, \qquad
    \transq[n]{q} = \rotq[n]{1}\,\reltrans{q} + \transq[n]{1},
    \label{eq:relative-pose}
\end{equation}
where $\relrot{q} = {\rotq[1]{1}}^{\!\top}\,\rotq[1]{q}$ 
and $\reltrans{q} = {\rotq[1]{1}}^{\!\top}(\depthscalar{q}\transq[1]{q} - \transq[1]{1})$
describe that rigid transformation.
Note that all quantities in Equation~\ref{eq:relative-pose} except the depth scalars and the poses of the first proxy are known during initialization (Section~\ref{sec:proxy-generation}). The resulting bundle-adjustment objective takes a similar form to Eq.~\ref{eq:smoothness}, but the reprojection error is summed over all $\nprompts$ proxies using the poses of Equation~\ref{eq:relative-pose} and the depth scalars $\depthscalar{q}$
 (see Supp.\ Section~\ref{supp:ba} for the full objective). 
We optimize the 
$6(\nframes{-}1) + (\nprompts{-}1)$ free variables
with 
Levenberg--Marquardt, initializing poses from Algorithm~\ref{supp:alg:tracking} with all $\depthscalar{q} = 1$.

\vspace{-0.5em}
\subsection{Implementation Details}
\label{sec:implementation}
\vspace{-0.5em}

\paragraph{Synthetic dataset.}
We construct a synthetic training set of $35{,}000$ paired (source, proxy) video sequences using 3D assets by Trellis-500K from Objaverse~\citep{xiang2025structured,deitke2023objaverse}.
Source videos are rendered in Blender~\citep{blender} following the composition-rendering protocol of recent diffusion-rendering work~\citep{liang2025diffusionrenderer,zhang2025unilight,liang2026luxdit}, with randomized objects, scene composition, backgrounds, ground planes, camera trajectories, and per-object rigid motion.
For each rendered object we sample a random marked pixel inside its visibility mask and use the recorded per-frame, per-object \dof{6} motion to render the corresponding proxy video with PyTorch3D~\citep{ravi2020pytorch3d}.
Each proxy video preserves the motion of exactly one object from its paired source video.
Additional details about dataset generation are provided in Supp.\ Section~\ref{supp:data}.

\vspace{-0.5em}
\paragraph{Fine-tuning.}
We build on the Wan-14B backbone~\citep{wan2025}, attaching rank-$64$ LoRA adapters for parameter-efficient fine-tuning.
The noise schedule offset is set to $\anchoroffset = 500$ steps.
We use a learning rate of $2{\times}10^{-4}$ during low-resolution fine-tuning and $5{\times}10^{-5}$ during high-resolution fine-tuning, and adopt the standard Wan flow-matching loss as implemented in DiffSynth~\citep{diffsynth}.

Fine-tuning proceeds in three stages totaling 100k iterations. The first stage trains for 80k iterations at $256{\times}256$ resolution with 29 frames. The second stage trains for 10k iterations at $512{\times}512$ resolution with 29 frames. The third stage trains for a further 10k iterations at $512{\times}512$ resolution with 49 frames. All training was conducted on $4{\times}$NVIDIA H100 GPUs, with per-stage wall-clock times of approximately 2.5, 1, and 2 days, respectively, for a total of approximately 22 GPU days. At inference, proxy video generation with $50$ flow-matching denoising steps requires 5.5 minutes on a single NVIDIA H100 GPU.

\vspace{-0.5em}
\paragraph{Classifier-free guidance.}
With probability $0.15$ during training we drop the conditioning by zeroing the source latents and noising the first proxy frame at the global timestep $\timestep$ rather than the anchor timestep $\anchortimestep$ (i.e., disabling the anchor schedule for the dropped sample).
The remaining $85\%$ of samples follow the regular conditioning and schedule.

\vspace{-0.5em}
\paragraph{Inference.} We first extract a $512 \times 512$ square crop from the input video, choosing the largest crop that keeps the query point as close to the center as possible. When the focal length is not known, we either set it to a value corresponding to a $45$-degree field of view or use Depth Anything 3~\citep{lin2025depth} to estimate it from the input video sequence---both approaches work well in practice. Figure~\ref{fig:teaser} uses the fixed field of view, and we assess sensitivity to the choice of focal length in Supp.\ Section~\ref{supp:results}.

\vspace{-0.5em}
\section{Evaluation}
\label{sec:evaluation}
\vspace{-0.5em}
 
We evaluate \method{} against a diverse set of baselines on two established benchmarks for object pose tracking in dynamic scenes, on our simulated dataset, and on challenging in-the-wild scenes.
We demonstrate state-of-the-art quantitative results on datasets where ground-truth is available, as well as qualitative results showing our method extends to challenging situations (e.g., textureless, shiny, transparent, or non-rigid surfaces) where other methods fail.

\vspace{-0.75em}
\paragraph{Datasets.} We evaluate on dynamic pose estimation datasets HO3D~\citep{hampali2020honnotate} (13 sequences of hand--object manipulation with dynamic occlusions), YCBInEOAT~\citep{wen2020se} (9 sequences of dual-arm robotic manipulation), and a held-out set of 14 sequences from our synthetic dataset (Section~\ref{sec:implementation}).
As these datasets involve tracking rigid objects (and our approach handles more general scenes), we also provide qualitative results on challenging in-the-wild videos.
 
All predicted poses are expressed relative to the ground truth pose in the first frame provided at a query point placed randomly on the object's visible surface.
To resolve the inherent scale ambiguity of monocular methods, every method's predicted depth is scaled to be consistent with the ground-truth depth at the first frame.
All baselines use the same input frames for fair comparison.
 Further dataset details are given in Supp.\ Section~\ref{supp:datasets}.

\vspace{-0.75em}
\paragraph{Baselines.} We compare against model-based and model-free pose estimators---FoundationPose~\citep{wen2024foundationpose} (in both registration and tracking modes), Any6D~\citep{lee2025any6d}, One2Any~\citep{liu2025one2any}, Oryon~\citep{corsetti2024oryon}, ConceptPose~\citep{kuang2026conceptpose}, and BundleSDF~\citep{wen2023bundlesdf}---as well as custom 6-DoF tracking pipelines we built from a 3D point tracker based on Spatial Tracker V2~\citep{xiao2024spatialtracker} and a tracker based on CoTracker 3~\citep{karaev2025cotracker3} with Depth Anything 3~\citep{lin2025depth}. 
For the custom trackers, we recover 6-DoF information using 3D tracking information from pixel neighborhoods (\textit{pixel level}) or from an entire object using ground-truth object segmentation masks (\textit{object level}).
See Supp.\ Section~\ref{supp:baselines} for a detailed description of baselines.

\begin{figure*}[t!]
    \begin{center}
    \includegraphics[width=1\textwidth]{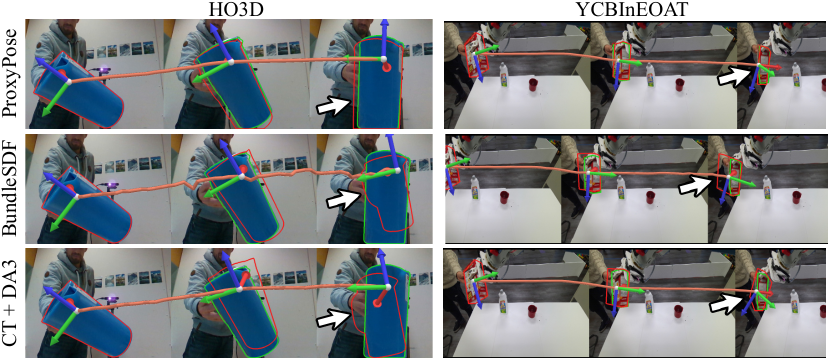}
    \end{center}
    \vspace{-1em}
   \captionof{figure}{Results on HO3D and YCBInEOAT. For each method, we show the tracked point (white dot), estimated orientation (coordinate axes), and trajectory, where we add a horizontal offset to connect the frames for visualization (pink line). Ground-truth and predicted object poses are shown as green and red contours, respectively. \method{} (one query) produces accurate and temporally smooth pose estimates from monocular video alone, while BundleSDF~\citep{wen2023bundlesdf} and CoTracker 3 + Depth Anything 3 (pixel level) require depth as input and exhibit drift and jitter (arrows).} 
    \label{fig:qualitative-results}

\captionof{table}{Quantitative evaluation on HO3D and YCBInEOAT. All poses are relative to the first frame, with scale aligned at that frame. \textbf{Bold}: best; \underline{underline}: second best (lower is better for all metrics). Methods are grouped by input requirements, from most supervision (top) to least (bottom). \method{} (one query) is the only method that requires no 3D model, depth, or object mask.}
\label{tab:quantitative-results}
\footnotesize
\setlength{\tabcolsep}{2.4pt} 
\resizebox{\textwidth}{!}{
\begin{tabular}{l|ccc|ccccc|ccccc}
\toprule
& \multicolumn{3}{c|}{Addl.\ Inputs} & \multicolumn{10}{c}{Evaluation Benchmarks} \\
\cmidrule(lr){2-4} \cmidrule(lr){5-14}
Method
& \vheader{3D Model}
& \vheader{Depth*}
& \vheader{Obj.\ Mask$\dagger$}
& \multicolumn{5}{c|}{HO3D}
& \multicolumn{5}{c}{YCBInEOAT}\\[-15pt]
\cmidrule(lr){5-9} \cmidrule(lr){10-14}
& & &
& \metric{ATE}{mm}
& \metric{ARE}{deg}
& \metric{RPE-t}{mm}
& \metric{RPE-r}{deg}
& \metric{2D-dist}{px}
& \metric{ATE}{mm}
& \metric{ARE}{deg}
& \metric{RPE-t}{mm}
& \metric{RPE-r}{deg}
& \metric{2D-dist}{px}\\
\midrule
FoundationPose (registration)
& \cb & \cb & & 53.17 & 74.55 & 37.09 & 63.36 & 37.09 & 95.84 & 118.3 & 47.35 & 71.12 & 31.19 \\
FoundationPose (track)
& \cb & \cb & & 25.44 & 15.38 & 6.156 & 2.874 & 18.05 & 50.19 & 18.58 & 11.17 & 3.579 & 10.97 \\
Oryon
& & \cb & & 40.42 & 37.97 & 23.52 & 23.70 & 45.69 & 41.17 & 41.45 & 20.88 & 34.07 & 25.63\\
\midrule
Any6D
& & \cb & \cb & 64.63 & 76.18 & 41.08 & 48.85 & 56.23 & 86.26 & 67.46 & 35.70 & 37.38 & 37.53 \\
One2Any
& & \cb & \cb & 52.02 & 88.99 & 12.86 & 17.53 & 55.31 & 35.31 & 33.99 & 10.78 & 10.29 & 23.33\\
ConceptPose
& & \cb & \cb & 34.24 & 79.95 & 38.89 & 94.89 & 38.14 & 43.04 & 68.38 & 37.52 & 75.67 & 26.42\\
CoTracker 3 + DA3 (object level)
& & \cb & \cb & 25.02 & 19.56 & 6.630 & 2.927 & 14.69 & 44.80 & 17.12 & 10.79 & 4.940 & 7.836\\
SpatialTrackerV2 (object level)
& & \cb & \cb & 25.87 & 24.27 & 1.867 & 1.477 & 25.09 & 32.57 & 22.35 & 5.706 & 3.227 & 8.411 \\
BundleSDF
& & \cb & \cb & 23.24 & 14.28 & 5.949 & 2.706 & 14.92 & 42.06 & 13.88 & 10.34 & 4.255 & \textbf{6.792} \\
\midrule

SpatialTrackerV2 (pixel level)
& & \cb & & 26.14 & 22.16 & 1.898 & 3.042 & 25.89 & 33.01 & 23.30 & 6.441 & 8.496 & 7.487\\
CoTracker 3 + DA3 (pixel level)
& & \cb & & 26.62 & 17.85 & 5.596 & 4.424 & \underline{12.07} & 47.65 & 24.56 & 10.67 & 6.503 & \underline{7.003}\\
\midrule 
ProxyPose (one query)
& & & & \underline{15.79} & 5.126 & \textbf{1.193} & 0.9768 & \textbf{8.016} & \underline{30.07} & 15.07 & \textbf{4.330} & 2.630 & 8.496\\
ProxyPose (two queries)
& & & \cb & \textbf{15.42} & \textbf{3.941} & \underline{1.220} & \underline{0.8523} & 12.31 & 31.62 & \underline{7.764} & 4.820 & \underline{2.269} & 7.969\\
ProxyPose (three queries)
& & & \cb & 18.55 & \underline{4.297} & 1.309 & \textbf{0.8206} & 14.63 & \textbf{26.32} & \textbf{6.476} & \underline{4.435} & \textbf{2.153} & 7.770\\
\bottomrule
\end{tabular}
}
\vspace{2pt}
{\scriptsize *We use DA3 to provide depth inputs.}
{\scriptsize $\dagger$ Object masks required in one or more video frames.}
\vspace{-2em}
\end{figure*}

As summarized in the ``Additional Input'' columns of Table~\ref{tab:quantitative-results}, all baselines require at least one of a 3D model, depth input, or object mask (we provide depth inputs using Depth Anything 3~\citep{lin2025depth}).
We evaluate \method{} variants using one, two, and three query pixels.
The multi-query configurations fuse proxy videos via bundle adjustment (Section~\ref{sec:rigidity}) and incorporate an object mask in the first frame to place all queries on the same object.
\method{} (one query) is the only method that operates from monocular RGB alone. 
All methods that do not use a 3D model output the 6-DoF pose relative to the first frame.

 \vspace{-0.75em}
\paragraph{Metrics.} The evaluation metrics are computed on poses relative to the first frame.
We report absolute translation error (ATE) and absolute rotation error (ARE), as well as relative pose errors in translation (RPE-t) and rotation (RPE-r)~\citep{sturm2012benchmark}.
We additionally report 2D reprojection distance (2D-dist) as an image-space measure, which is the mean pixel distance between predicted and ground-truth tracked points.

\subsection{Results}
\label{sec:results}
Quantitative results are presented in Table~\ref{tab:quantitative-results} and qualitative comparisons in Figure~\ref{fig:qualitative-results}.
\method{} achieves substantially lower rotation and translation errors than all baselines on both benchmarks, and is noticeably more stable and consistent with the ground-truth pose in Figure~\ref{fig:qualitative-results}. 
Since the objects in HO3D and YCBInEOAT are rigid, multi-query bundle adjustment can be applied, improving performance in most cases.
Even without rigidity constraints or bundle adjustment, \method{} (one query) achieves state-of-the-art performance. 
See Supp.\ Tables~\ref{supp:tab:extra-metrics}--\ref{supp:tab:synthetic-results} for per-sequence results and quantitative results on the synthetic dataset, as well as Table~\ref{tab:std-quantitative-results}, an extended version of Table~\ref{tab:quantitative-results} with standard deviation statistics.

\vspace{-0.75em}  
\paragraph{In-the-wild results.}

\begin{figure*}[t]
    \centering
    \includegraphics[width=0.98\textwidth]{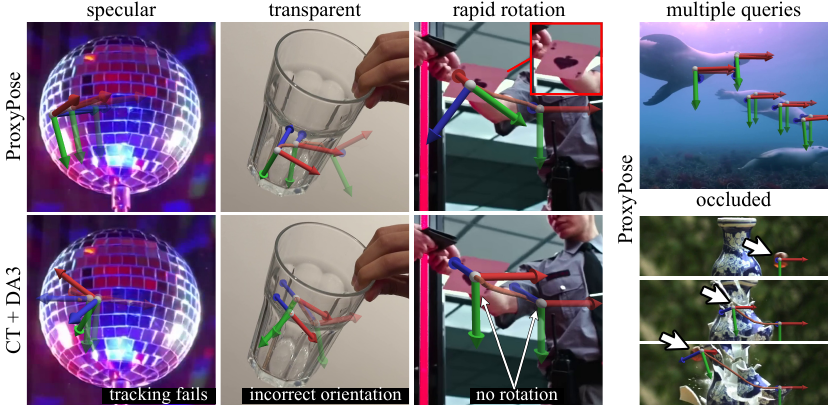}
    \vspace{-0.5em}
   \caption{Challenging in-the-wild videos. \method{} successfully tracks surface regions on a specular disco ball, transparent glass, and a spinning card---scenes where CoTracker 3 + Depth Anything 3 fails~\citep{karaev2025cotracker3,lin2025depth}---on seals spinning underwater, and it can follow objects through occlusions. \textit{See the \webpage{} for animated results and additional comparisons.}}
    \label{fig:in-the-wild}
    \vspace{-1em}
\end{figure*}

Figure~\ref{fig:teaser} and Figure~\ref{fig:in-the-wild} show \method{} applied to challenging internet videos featuring fast motion, occlusions, non-rigid objects, cartoon scenes, specular or transparent surfaces, and tracking of multiple deforming regions.
These scenes are beyond the capabilities of most baselines, including Cotracker 3 + Depth Anything 3, which produces degraded or failed tracks (see \webpage{}), while \method{} produces plausible 6-DoF trajectories.
 
\vspace{-0.75em}  
\paragraph{Ablation study.} In Table~\ref{tab:ablation}, we ablate LoRA rank (32, 64, 128), training set size (300, 3k, 35k sequences), and use of the noise schedule offset against our default configuration (Wan-14B, rank-64, 35k sequences) on a held-out set of our synthetic dataset.
Our proposed configuration achieves the best performance.
Interestingly, a dataset size of only 300 samples already achieves compelling performance; at such scales it may be possible to perform fine-tuning on hand-curated datasets that are specialized to a particular task.

\begin{wraptable}[9]{r}{0.5\textwidth}
\vspace{-1em}
\caption{Ablation study on the synthetic dataset.}
\label{tab:ablation}
\vspace{-0.5em}
\centering
\resizebox{\linewidth}{!}{
\begin{tabular}{lcccccc}
\toprule

Ablation
& \metric{ATE}{mm}
& \metric{ARE}{deg}
& \metric{RPE-t}{mm}
& \metric{RPE-r}{deg}
& \metric{2D-dist}{px} \\

\midrule

LoRA Rank 32
& 94.08 & 56.15 & 11.99 & 6.041 & 149.9\\

LoRA Rank 128
& 88.21 & 44.76 & 10.44 & 4.602 & 141.6\\

\midrule

Dataset Size 300
& 88.71 & 46.08 & 9.079 & 4.544 & 138.9\\

Dataset Size 3K
& 87.70 & 43.88 & 9.067 & 5.057 & 131.4\\

\midrule

Noise Offset $0$
& 507.5 & 98.48 & 354.5 & 41.76 & 1009 \\

\midrule

Full Model
& \textbf{81.65} & \textbf{36.83} & \textbf{8.181} & \textbf{3.178} & \textbf{117.3} \\

\bottomrule
\end{tabular}}
\end{wraptable}

\begin{figure*}[t]
    \centering
    \includegraphics[width=\textwidth]{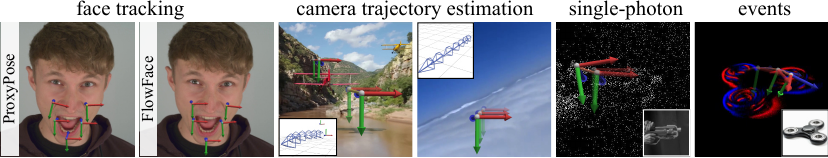}
    \vspace{-1.5em}
    \caption{Additional applications. Face tracking compared to FlowFace~\citep{taubner20243d}: \method{} recovers smooth \dof{6} face pose without a face-specific model. Camera trajectory estimation: from just one or two query pixels on background scenery or on a cloud, \method{} tracks the camera's motion (COLMAP~\citep{schonberger2016structure} fails for the cloud scene). Zero-shot application to alternative capture modalities: single-photon camera footage of a NeRF gun~\citep{xie2026inter} and event camera data of a hand spinner~\citep{prophesee2024recordings} (intensity images shown in insets). }
    \label{fig:applications}
    \vspace{-1em}
\end{figure*}

\paragraph{Additional applications.} We explore several other applications of \method{} in Figure~\ref{fig:applications}.
Placing a query pixel on a highly non-rigid face yields 6-DoF head pose trajectories comparable to FlowFace without a face-specific model.
Aggregating estimates from multiple query points on static background surfaces recovers camera-pose trajectories, succeeding on a sequence where COLMAP fails due to insufficient feature matches.
Finally, \method{} generalizes to footage from event cameras and single-photon cameras, suggesting that the learned motion priors may transfer across imaging modalities.
 
\section{Discussion and Conclusion}
\label{sec:discussion}
\vspace{-0.5em}
\begin{wrapfigure}[20]{r}{55.9mm}
\vspace{-4em}
    \centering
    \includegraphics{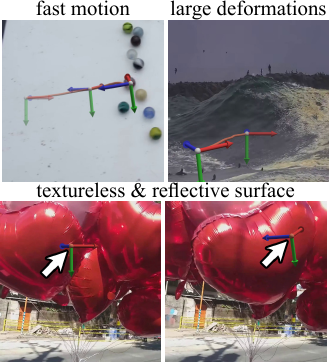}
    \caption{Limitations. Fast motion (marbles) can result in blur from the VAE, degrading pose recovery. Tracking can also drift for fluid surfaces where locally rigid motion is ill-defined (waves) or for textureless/reflective objects (balloons).}
    \label{fig:limitations}
\end{wrapfigure}

Despite its generality, \method{} inherits limitations from the underlying video model and proxy-based formulation (Figure~\ref{fig:limitations}).
Fast motion can exceed the VAE's encoding capabilities, producing blurred proxy frames that degrade contour detection.
The approximate local rigidity implied by the proxy can be inconsistent with scenes where tracking is ill-defined (e.g., surface regions on a fluid), and we sometimes observe pose drift for textureless or reflective objects under complex motion.
Finally, inference requires multiple minutes on a high-end GPU; leveraging efficient autoregressive video models~\citep{yin2025slow} to enable real-time tracking is a promising direction for future work.

Our results suggest that large video generation models can serve as general-purpose backbones for 3D motion understanding, complementing or even replacing task-specific foundation models for perception.
More broadly, our use of video-to-video translation to transform a difficult perceptual problem into one amenable to classical solvers could extend to other tasks such as articulated body tracking, non-rigid surface reconstruction, or dense scene flow estimation.
\section*{Acknowledgments}
\label{sec:acknowledgments}
\vspace{-0.5em}

DBL and KNK acknowledge support of NSERC under the RGPIN program. DBL also acknowledges support from the Canada Foundation for Innovation, the Ontario Research Fund, and the Digital Research Alliance of Canada.

{
    \small
    \bibliographystyle{ieeenat_fullname}
    \bibliography{ref}
}

\newpage

\setcounter{section}{0}
\setcounter{figure}{0}
\setcounter{table}{0}
\setcounter{equation}{0}
\setcounter{algorithm}{0}

\renewcommand{\thesection}{\Alph{section}}
\renewcommand{\thefigure}{S\arabic{figure}}
\renewcommand{\thetable}{S\arabic{table}}
\renewcommand{\theequation}{S\arabic{equation}}
\renewcommand{\thealgorithm}{S\arabic{algorithm}}

\begin{center}
    {\Large\bfseries ProxyPose: 6-DoF Pose Tracking\\
    via Video-to-Video Translation\\[0.75em]}
    {\large Supplementary Material\\[1em]}
\end{center}

\vspace{1em}

\renewcommand{\contentsname}{Supplementary Table of Contents}
\startcontents[supp]
\printcontents[supp]{}{1}{\setcounter{tocdepth}{2}}

\vspace{1.5em}

\paragraph{Broader Impact.}
Our work could democratize 6-DoF tracking by removing the need for CAD models, depth sensors, or large task-specific datasets, making robust tracking more accessible for applications in robotics, augmented reality, and scientific video analysis. On the other hand, improved tracking capabilities carry inherent dual-use risks. More robust pose estimation from monocular video could facilitate surveillance applications or enable more capable autonomous systems in sensitive contexts. We also note that our reliance on a pretrained video diffusion model means we inherit any biases present in that model's training data, which may affect tracking robustness across different visual domains. We encourage the community to consider these factors when building on this work.

\section{Additional Implementation Details}
\label{supp:implementation}

\subsection{Proxy Video Generation}
\label{supp:generation}

\paragraph{Token concatenation and 4D RoPE.} We extend the pretrained 3D rotary position embedding (RoPE) to encode a stream identifier that distinguishes the source and proxy tokens. The backbone model partitions each attention head ($d_{\text{head}}{=}128$ real dimensions) into three axes (temporal: $d_t{=}44$, height: $d_h{=}42$, and width: $d_w{=}42$) with each using the standard frequency schedule $\omega_k = \theta^{-2k/d_{\mathrm{axis}}}$ with $\theta=10,000$. Here, we re-purpose the last complex component of the width axis as a binary stream discriminator. For source tokens, the frequency entry retains its standard positional value, while for proxy tokens, it is overwritten with $-1 = e^{i\pi}$. This introduces a $\pi$ phase flip in one frequency slot, causing cross-stream query--key dot products to differ from within-stream products and enabling the attention layers to distinguish the two token streams without adding significant additional parameters. 

\paragraph{LoRA configuration.}

We insert low-rank (LoRA) adapters of rank $r=64$ into the $\mathbf{Q}$, $\mathbf{K}$, $\mathbf{V}$, and output projections of every self- and cross-attention block, as well as the two linear layers of each feed-forward network (\texttt{ffn.0}, \texttt{ffn.2}). The scaling factor is $\alpha{=}r{=}64$ and no dropout is applied. In total, it yields ${\sim}307\text{M}$ trainable parameters (${\sim}2.2\%$ of the frozen Wan-14B backbone), distributed across 800 adapter weight matrices. 

\paragraph{Text conditioning.}
The fixed text prompt supplied to the cross-attention layers is held constant
across training and inference. 

\begin{quote}\small\itshape
``A perfectly geometric cube rotating slowly in place against a pure black
background.\ The center of the cube is rigidly attached to the surface of the
object it is tracking.\ The cube has six different solid-colored faces: red,
green, blue, yellow, white, and cyan.\ Each color is assigned to one face and
remains permanently attached to that same face throughout the entire video as
the cube rotates.\ Sharp edges, flat faces, no texture, no gradients, no
reflections, no shadows, no extra objects, static camera, high contrast.''
\end{quote}

\paragraph{Noise schedule offset details.}
The noise schedule offset is $\anchoroffset = 500$ steps on the Wan
flow-matching schedule. When $\timestep < \anchoroffset$, the first-frame
proxy tokens are kept at the data manifold ($\anchortimestep = 0$), so the
model effectively receives the clean first frame only on the lowest-noise
steps. We provide ablation results for different noise offsets, $\anchoroffset \in \{0, 500\}$, in Section~\ref{sec:results}.

\subsection{Proxy Video Tracking}
\label{supp:tracking}

Algorithm~\ref{supp:alg:tracking} provides a high-level summary of the proxy video tracking procedure, and Algorithm~\ref{supp:alg:tracking-detailed} gives the full pseudocode including the correspondence-matching subroutine. Below we describe each component in detail.

\paragraph{Choice of proxy geometry.} We use a cube as the proxy primitive because it provides sufficient geometric constraints for robust pose recovery while remaining simple enough to detect and track reliably. When initialized with its $[1,1,1]^\top$ diagonal aligned to the camera ray, exactly three faces are visible with approximately equal projected area, yielding up to seven non-coplanar 3D--2D vertex correspondences---above the four required by PnP. Each visible face projects to a quadrilateral that can be robustly detected with standard contour-fitting routines. Simpler primitives such as tetrahedra typically expose only one or two small triangular faces from any given viewpoint, providing insufficient correspondences for reliable pose estimation, while more complex polyhedra produce smaller faces that are more difficult to segment. The cube is therefore a simple convex solid that reliably presents three large, well-separated quadrilateral faces from a single viewpoint, making it straightforward to track with classical image-processing pipelines.

\begin{algorithm}[t]
\caption{Proxy video tracking.}
\label{supp:alg:tracking}
\begin{algorithmic}[1]
\Require Proxy video $\proxyhat = \{\proxyhat[n]\}_{n=1}^{\nframes}$, intrinsics $\intrinsics$, cube vertices $\cubecornersset$, face colors $\{c_i\}_{i=1}^{6}$, initial pose $(\rot_1, \trans_1)$
\Output Per-frame poses $\{(\rotf,\, \transf)\}_{n=1}^{\nframes}$ and 2D--3D correspondences $\{\corrset\}_{n=1}^{\nframes}$
\For{$n = 1, \dots, \nframes$}
    \item[] \vspace{-0.7em}
    \State \textit{// detect cube faces and image coordinates of their corners}
    \State Create segmentation masks from $\proxyhat[n]$  based on the known cube face colors $\{c_i\}_{i=1}^{6}$
    \State Detect quadrilateral contours from masks and extract sub-pixel corners $\{\imgpoint_{i,k}\}$ 
    \item[] \vspace{-0.7em}
    \State \textit{// initialize pose estimate}
    \If{$n = 1$}
        \State $(\posepriorrot,\, \posepriortrans) \leftarrow (\rot_1, \trans_1)$
    \EndIf
    \item[] \vspace{-0.7em}
    \State \textit{// refine pose using all visible faces}
    \State $\corrset\leftarrow$ Match each detected corner $\imgpoint_{i,k}$ to the corresponding cube vertex under $(\posepriorrot,\, \posepriortrans)$
    \State $(\rotf,\, \transf) \leftarrow \texttt{solvePnP}$ over all matched 2D--3D correspondences
    \State $(\posepriorrot, \posepriortrans) \leftarrow$ constant-velocity extrapolation of $(\rotf,\, \transf)$
\EndFor
\State \Return $\{(\rotf,\, \transf, \corrset)\}_{n=1}^{\nframes}$
\end{algorithmic}
\end{algorithm}

\paragraph{Color thresholds.}
Each cube face is assigned a saturated, well-separated color in RGB space. The colors are listed below in BGR format (following the OpenCV convention):

\begin{quote}\small\itshape
\begin{center}
``Front: white, $(255, 255, 255)$.\\
Back: yellow $(0, 255, 255)$.\\
Left: cyan $(230, 230, 0)$.\\
Right: green $(0, 240, 75)$.\\
Bottom: blue $(255, 30, 30)$.\\
Top: red $(10, 10, 255)$.'' 
\end{center}
\end{quote}
We compute per-face masks by measuring the Euclidean distance in RGB space between each pixel and the target face color. Pixels with a distance below a threshold of 70 are assigned to the corresponding cube face.

\paragraph{Contour and corner refinement.}
Quadrilateral candidates are extracted using \texttt{cv2.findContours} followed by \texttt{cv2.approxPolyDP} with approximation tolerance $\varepsilon = 0.02 \cdot \mathrm{perimeter}$. We retain only $4$-vertex contours whose area exceeds a fixed minimum threshold of 500 pixels, chosen to reject contours smaller than the expected projected face area at the maximum tracking depth. Corner locations are refined using
\texttt{cv2.cornerSubPix} with a $5{\times}5$ search window and a termination criterion of either $100$ iterations or sub-pixel change below $10^{-3}$.

\paragraph{PnP and correspondence matching.}
We use \texttt{cv2.solvePnP} with the IPPE-Square solver for both single-face initialization and the multi-correspondence refinement. 

Face color is not used for correspondence matching, as it is not sufficiently reliable. Instead, each detected quadrilateral contour $C = (\imgpoint_1, \dots, \imgpoint_4)$ is greedily matched to the closest reprojected cube face by minimizing the maximum corner reprojection error across all four vertices. Formally, the distance between contour $C$ and candidate face $i$ is
\[
d(C, i) =
\min_{r \in \{1,\dots,4\}}
\max_{k \in \{1,\dots,4\}} \left\lVert
\proj(\cubecorner[\facevertices{i}[\sigma_r(k)]]) - \imgpoint_k
\right\rVert_2 ,
\]
where $\facevertices{i}$ denotes the four vertex indices of face $i$, $\cubecorner[\facevertices{i}[k]]$ is the $k$-th 3D vertex of that face, $\proj$ denotes perspective projection under the current pose estimate, and $\sigma_r$ enumerates the four cyclic permutations of $(1,2,3,4)$ (see Algorithm~\ref{supp:alg:tracking-detailed}). Each contour is assigned to the face $i$ that minimizes $d(C, i)$. After all visible contours are assigned, we perform an additional multi-correspondence PnP refinement step.

\begin{algorithm}[t]
\caption{Proxy video tracking (detailed pseudocode).}
\label{supp:alg:tracking-detailed}
\begin{algorithmic}[1]
\Require Proxy video $\proxyhat = \{\proxyhat[n]\}_{n=1}^{\nframes}$, intrinsics $\intrinsics$, cube vertices $\cubecornersset$, face colors $\{c_i\}_{i=1}^{6}$, initial cube pose $(\rot_1, \trans_1)$
\Output Per-frame poses and 2D--3D correspondences $\{(\rotf,\, \transf,\, \corrset)\}_{n=1}^{\nframes}$
\State Let $\facevertices{i} \subset \{1,\dots,8\}$ denote the four vertex indices of cube face $i$
\State $(\posepriorrot,\, \posepriortrans) \leftarrow \texttt{None}$; \; $\mathbf{v}_{\trans} \leftarrow \mathbf{0}$; \; $\mathbf{v}_{\rot} \leftarrow \mathbf{0}$
\For{$n = 1, \dots, \nframes$}
    \State \textit{// detect cube faces and corresponding 2D corner coordinates}
    \State $\mathcal{Q} \leftarrow \emptyset$ \Comment{set of detected cube face indices}
    \For{$i = 1, \dots, 6$}
        \State $M_i \leftarrow \texttt{createMask}(\proxyhat[n],\; c_i)$ \Comment{color mask for face $i$}
        \State $\{\gamma_m\} \leftarrow \texttt{findContours}(M_i)$
        \State $\{\hat{\gamma}_m\} \leftarrow \texttt{approxQuad}(\{\gamma_m\})$ \Comment{fit quadrilateral to cube face contour}
        \If{$|\hat{\gamma}_m| > 0$}
            \State $(\imgpoint_{i,1},\dots,\imgpoint_{i,4}) \leftarrow \texttt{cornerSubPix}(\hat{\gamma}_m)$ \Comment{sub-pixel refinement of quad corners}
            \State $\mathcal{Q} \leftarrow \mathcal{Q} \cup \{i\}$
        \EndIf
    \EndFor
    \State
    \State \textit{// initialize pose estimate}
    \If{$(\posepriorrot,\, \posepriortrans) = \texttt{None}$} \Comment{initialize first frame pose }
        \State $(\posepriorrot,\, \posepriortrans) \leftarrow (\rot_1, \trans_1)$
    \EndIf
    \State
    \State \textit{// match detected corners to reprojected cube vertices of each face, and solve for pose}
    \State $\corrset \leftarrow \Call{findCorrespondences}{\posepriorrot,\, \posepriortrans,\, \mathcal{Q},\, \{\imgpoint_{i,k}\}}$
    \State $(\rotf,\, \transf) \leftarrow \texttt{solvePnP}(\corrset,\; \intrinsics)$ \Comment{refine pose with all correspondences}
    \State $\mathbf{v}_{\trans} \leftarrow \transf - \trans_{n-1}$; \; $\mathbf{v}_{\rot} \leftarrow \log(\rotf \rot_{n-1}^{\top})$ \Comment{update velocity}
    \State $(\posepriorrot,\, \posepriortrans) \leftarrow (\exp(\mathbf{v}_{\rot})\,\rotf,\;\, \transf + \mathbf{v}_{\trans})$ \Comment{estimate pose of next frame assuming constant velocity}
\EndFor
\State
\State \Return $\{(\rotf,\, \transf,\, \corrset)\}_{n=1}^{\nframes}$
\State
\State \textit{// match cube face corner 2D coordinates to 3D cube vertices}
\Function{findCorrespondences}{$\rot,\, \trans,\, \mathcal{Q},\, \{\imgpoint_{i,k}\}$}
    \State $\corrset \leftarrow \emptyset$
    \State Let $\sigma_r$ for $r = 1,\dots,4$ denote the four cyclic permutations of $(1,2,3,4)$
    \For{each detected contour corner $(\imgpoint_{i,1},\dots,\imgpoint_{i,4})$}
        \State $d^{*} \leftarrow \infty$; \; $i^{*} \leftarrow \texttt{None}$; \; $r^{*} \leftarrow \texttt{None}$
        \For{$i \in 1,\ldots,6$} \Comment{iterate over candidate cube faces}
            \State $\{\hat{\imgpoint}_k\}_{k \in \facevertices{i}} \leftarrow \proj(\intrinsics,\, \rot,\, \trans;\, \{\cubecorner[k]\}_{k \in \facevertices{i}})$
            \For{$r = 1, \dots, 4$} \Comment{iterate over cyclic orientations}
                \State $d \leftarrow \max_{k \in \{1,\dots,4\}} \|\hat{\imgpoint}_{\facevertices{i}[\sigma_r(k)]} - \imgpoint_{i,k}\|$
                \If{$d < d^{*}$}
                    \State $d^{*} \leftarrow d$; \; $i^{*} \leftarrow i$; \; $r^{*} \leftarrow r$
                \EndIf
            \EndFor
        \EndFor
        \For{$k = 1, \dots, 4$}
            \State $\corrset \leftarrow \corrset \cup \{(\cubecorner[\facevertices{i^{*}}[\sigma_{r^{*}}(k)]],\, \imgpoint_{i,k})\}$
        \EndFor
    \EndFor
    \State \Return $\corrset$
\EndFunction
\end{algorithmic}
\end{algorithm}

\paragraph{Constant-velocity propagation.}
Between frames, we propagate the pose with a constant-velocity model on $\mathfrak{se}(3)$. Translational velocity is estimated from the previous inter-frame translation, while rotational velocity is computed as the logarithm of the relative inter-frame rotation. We set the initial velocity to zero. If tracking fails for a frame (i.e. no valid quadrilateral is detected), we fall back to pure constant-velocity prediction and resume contour-based detection on the following frame.

\subsection{Bundle Adjustment}
\label{supp:ba}

\paragraph{Objective.}
The bundle-adjustment objective extends Equation~\ref{eq:smoothness} to sum over all proxies, with the depth scalar entering through the projection argument:
\begin{equation}
\resizebox{\textwidth}{!}{$\begin{aligned}
    \baLoss \;=\;
    \sum_{q=1}^{\nprompts} \sum_{n=1}^{\nframes}
        \!\sum_{(\imgpoint,\, \cubecorner[]) \in \corrsetq[n]{q}}\!
        \big\lVert \proj\!\big(\rotq[n]{q}\,\depthscalar{q}\cubecorner[] + \transq[n]{q}\big) - \imgpoint \big\rVert_{2}^{2}
    \;+\;
    \sum_{n=1}^{\nframes-1} \!\Big[
        \wtrans \big\lVert \transq[n+1]{1} \!-\! \transq[n]{1} \big\rVert_{2}^{2}
        + \wrot \big\lVert \log\!\big( \rotq[n+1]{1} \rotq[n]{1}{}^{\!\top} \big) \big\rVert_{\text{F}}^{2}
    \Big],
\end{aligned}$}
\label{eq:ba}
\end{equation}
where $\rotq[n]{q}$ and $\transq[n]{q}$ are given by Equation~\ref{eq:relative-pose}.
The free variables are the poses of proxy~1 at frames $n = 2,\ldots,\nframes$, together with the depth scalars $\{\depthscalar{q}\}_{q=2}^{\nprompts}$, giving $6(\nframes{-}1) + (\nprompts{-}1)$ degrees of freedom in total.
We minimize Equation~\ref{eq:ba} with Levenberg--Marquardt, initializing poses from Algorithm~\ref{supp:alg:tracking} and all $\depthscalar{q} = 1$.
When $\nprompts = 1$ this reduces to Equation~\ref{eq:smoothness}.

\paragraph{Solver.}
We optimize the reprojection residuals using Levenberg--Marquardt with analytic Jacobians and an L2 loss. We use the SciPy implementation of LM via \texttt{scipy.optimize.least\_squares}, retaining the default solver settings except for the robust loss specified in the main text. We allow a maximum of 2000 iterations, which in practice requires approximately 20 seconds for optimization over three proxies.

\paragraph{Velocity weights.}
We use $\wtrans = 200$ and $\wrot = 40$ in Equation~\ref{eq:ba}. 

\clearpage

\section{Synthetic Data Generation}
\label{supp:data}

\paragraph{Source video rendering.} 
We generate a collection of 35,000 source videos using Blender with a physically-based path-tracing renderer. Each video consists of 64 frames at $512 \times 512$ resolution with 64 samples per pixel. We source 3D assets from the Trellis-500K dataset. For each video, we sample the number of foreground objects from a uniform distribution over the set $\{1,\ldots,4\}$. Each object is independently rescaled and randomly rotated before being placed into the scene. Scene appearance is constructed using HDR environment maps that provide both illumination and background. To increase diversity, we apply random rotation, horizontal flipping with 50\% probability, and intensity scaling in the range $[0.7,1.4]$. A textured ground plane is added using randomly selected physically-based rendering materials, each applied with a random scale factor in $[1.5,2.5]$. We consider two motion regimes. In the \emph{drop} mode, objects are dropped from above the ground plane and simulated using rigid body dynamics under gravity, with randomized initial velocities, angular motion, restitution, and friction, while bounding walls keep objects in view. In the \emph{fly} mode, objects move in zero gravity with trajectories generated via Euler integration, enabling full control over motion. Objects are initialized within the camera frustum and assigned random linear and angular velocities. Videos are rendered using a pinhole camera model with fixed intrinsics and randomized extrinsics. The field of view is fixed at $45^\circ$, yielding a focal length $f_x = f_y \approx 618$ pixels for image width $w = 512$. The principal point is taken to be the image center, $c_x = c_y = 256$. The camera either remains fixed or follows a linear trajectory between two randomly sampled poses $A$ and $B$, parameterized by azimuth $\phi \in [0^\circ, 360^\circ]$, elevation $\theta \in [5^\circ, 60^\circ]$, and radius $r$. 

\paragraph{Proxy video rendering.}
Each proxy is rendered with PyTorch3D using the per-frame, per-object
\dof{6} transform extracted from the Blender scene. The proxy cube's
first-frame placement follows the same canonical rule as at inference (unit
distance, fixed screen size, corner aligned with the marked pixel), so the
training and inference distributions of first-frame proxies match exactly.

\paragraph{Train/validation/test split.}
From the full dataset, we hold out 100 sequences for validation. Additionally, we reserve 14 sequences to form a synthetic evaluation benchmark. The split is constructed to have no overlap with the training data in terms of 3D assets, backgrounds, ground planes, or motion trajectories.

\section{Evaluation Details}
\label{supp:evaluation}

\subsection{Datasets}
\label{supp:datasets}

We evaluate on HO3D~\citep{hampali2020honnotate}, which contains RGB-D sequences and ground-truth poses for hand--object interactions captured during continuous manual manipulation.
We use the HO3D\_v3 split, which comprises 13 video sequences featuring 4 YCB objects.
Unlike static tabletop setups, HO3D captures objects undergoing rotation and translation while being partially occluded by the manipulating hand, which requires tracking through dynamic occlusions.
 
We also evaluate on YCBInEOAT~\citep{wen2020se}, which was designed for evaluating 6-DoF tracking under robotic manipulation.
It contains 9 RGB-D video sequences of 5 YCB objects manipulated by a dual-arm robot across three task types: single-arm pick-and-place, within-hand manipulation, and pick-to-handoff transitions.
All sequences are captured by a mounted Azure Kinect sensor, which is used to provide precise ground-truth poses.

For each video sequence in HO3D and YCBInEOAT, we select a window of 49 contiguous frames that maximizes the rotation delta between the first and last frames, discarding any window whose first frame has less than 90\% object visibility. 
We select a query point on the tracked object by randomly selecting a pixel that is within the top 25\% of the pixels furthest away from the object boundary as given by the ground truth object mask in the first frame.
To compute the evaluation metrics, we compare the ground truth and predicted poses at the query point.

Finally, we evaluate on a held-out set from our own synthetic dataset (see Supp.\ Section~\ref{supp:data}).
We also provide qualitative results on a set of challenging in-the-wild examples from internet videos.

\subsection{Baseline Details} 
\label{supp:baselines}

We use publicly available codebases for all baselines except CoTracker 3 + Depth Anything 3, which we implement ourselves as described below.
\paragraph{FoundationPose~\citep{wen2024foundationpose}.} We evaluate FoundationPose in two modes. \emph{Track} uses a CAD model and ground-truth depth to perform frame-to-frame pose tracking initialized from the first frame. \emph{Registration} re-estimates the pose independently at each frame via model registration without temporal continuity. Both modes receive ground-truth depth maps.
\paragraph{Any6D~\citep{lee2025any6d}.} Any6D operates without a CAD model but requires depth and object masks as input.
\paragraph{One2Any~\citep{liu2025one2any} and Oryon~\citep{corsetti2024oryon}.} Both are model-free single-view matching methods that leverage depth and object masks to estimate pose from a single reference view.
\paragraph{ConceptPose~\citep{kuang2026conceptpose}.} ConceptPose uses vision-language reasoning with depth input for category-level pose estimation, removing the need for a CAD model but requiring depth.
\paragraph{BundleSDF~\citep{wen2023bundlesdf}.} BundleSDF performs joint tracking and neural implicit reconstruction from depth video without requiring a CAD model.
\paragraph{CoTracker 3 + Depth Anything 3~\citep{karaev2025cotracker3,lin2025depth}.} To assess whether dense 3D point tracking can substitute for explicit pose estimation, we implement two baselines using CoTracker 3~\citep{karaev2025cotracker3} (offline version) combined with monocular depth from Depth Anything V3 (DAV3)~\citep{lin2025depth}. The \emph{pixel-level} variant tracks an $11{\times}11$ neighborhood of pixels around the prompt point, lifts them to 3D using the estimated depth, and recovers the rigid-body transform via the Kabsch algorithm (SVD). The \emph{object-level} variant applies the same procedure to all tracked pixels within the ground-truth object mask, providing a stronger baseline that uses additional supervision. In both variants, the DAV3 depth map is globally scaled by a single multiplicative factor to match the ground-truth depth at the prompt pixel. When fewer than 3 tracked points remain visible (based on CoTracker's visibility mask), the last successfully estimated pose is held stationary.
\paragraph{SpatialTracker V2~\citep{xiao2024spatialtracker}.} SpatialTracker V2 extends point tracking to 3D using learned depth priors.

\clearpage
\section{Supplemental Results}
\label{supp:results}

It is encouraged to visit the \textbf{\webpage{}}, which contains video results and side-by-side comparisons for all sequences discussed in the main paper and supplement.

\paragraph{Additional metrics.}
In the main paper, we focus on pixel-level metrics that measure accuracy of the tracked local surface region. For completeness, we include additional object-level metrics (ADD and ADD-S~\citep{hinterstoisser2012model,xiang2018posecnn}) in Table~\ref{supp:tab:extra-metrics}.

\begin{figure*}[t!]
\captionof{table}{Quantitative evaluation on HO3D and YCBInEOAT. All poses are relative to the first frame, with scale aligned at that frame. \textbf{Bold}: best; \underline{underline}: second best.} 
\label{supp:tab:extra-metrics}
\centering
\footnotesize
\setlength{\tabcolsep}{2.4pt} 
\begin{tabular}{l|cc|cc}
\toprule
& \multicolumn{4}{c}{Evaluation Benchmarks} \\
\cmidrule(lr){2-5}
Method
& \multicolumn{2}{c|}{HO3D}
& \multicolumn{2}{c}{YCBInEOAT}\\
\cmidrule(lr){2-3} \cmidrule(lr){4-5}
& \metricup{ADD}{\%} & \metricup{ADD-S}{\%}
& \metricup{ADD}{\%} & \metricup{ADD-S}{\%} \\
\midrule
FoundationPose (registration)
& 49.2 & 85.2 & 29.0 & 76.1 \\
FoundationPose (track)
& 72.5 & 88.7 & 64.1 & 80.8 \\
Oryon
& 58.2 & 82.8 & 49.2 & 71.2 \\
\midrule
Any6D
& 42.5 & 75.4 & 38.5 & 62.2 \\
One2Any
& 39.5 & 85.4 & 53.0 & 75.8 \\
ConceptPose
& 41.4 & 74.3 & 41.9 & 69.2 \\
CoTracker 3 + DA3 (object level)
& 67.5 & 85.7 & 63.5 & 78.5 \\
SpatialTrackerV2 (object level)
& 71.6 & 89.0 & 59.7 & 81.0 \\
BundleSDF
& 75.2 & 89.4 & 64.7 & 79.3 \\
\midrule
SpatialTrackerV2 (pixel level)
& 71.5 & 89.2 & 59.5 & 80.9 \\
CoTracker 3 + DA3 (pixel level)
& 70.0 & 87.1 & 59.5 & 75.9 \\
\midrule 
ProxyPose (one query)
& \underline{82.5} & \underline{91.7} & \underline{72.4} & \underline{84.2} \\
ProxyPose (two queries)
& \textbf{83.8} & \textbf{92.4} & 70.3 & 84.1 \\
ProxyPose (three queries)
& 81.5 & 91.2 & \textbf{73.9} & \textbf{86.9} \\
\bottomrule
\end{tabular}
\end{figure*}

\paragraph{Per-sequence results on HO3D.}
\label{supp:per-sequence-ho3d}
Table~\ref{supp:tab:per-sequence-ho3d} reports per-sequence results on HO3D for \method{} (one query), FoundationPose (track), and CoTracker 3 + DA3 (pixel level). \method{} achieves the best mean performance across all five metrics, with particularly strong gains in rotation accuracy (ARE) and relative pose error (RPE-r).

\begin{table*}[t]
\centering
\caption{Per-sequence results on HO3D. All poses are relative to the first frame. \textbf{Bold}: best; \underline{underline}: second best.}
\label{supp:tab:per-sequence-ho3d}
\footnotesize
\setlength{\tabcolsep}{3pt}
\resizebox{\textwidth}{!}{
\begin{tabular}{l|ccccc|ccccc|ccccc}
\toprule
& \multicolumn{5}{c|}{ProxyPose (one query)} & \multicolumn{5}{c|}{FoundationPose (track)} & \multicolumn{5}{c}{CoTracker 3 + DA3 (pixel level)} \\
\cmidrule(lr){2-6} \cmidrule(lr){7-11} \cmidrule(lr){12-16}
Sequence
& \metric{ATE}{mm} & \metric{ARE}{deg} & \metric{RPE-t}{mm} & \metric{RPE-r}{deg} & \metric{2D-dist}{px}
& \metric{ATE}{mm} & \metric{ARE}{deg} & \metric{RPE-t}{mm} & \metric{RPE-r}{deg} & \metric{2D-dist}{px}
& \metric{ATE}{mm} & \metric{ARE}{deg} & \metric{RPE-t}{mm} & \metric{RPE-r}{deg} & \metric{2D-dist}{px} \\
\midrule
0000000 & 6.409 & 4.816 & 0.8187 & 0.8495 & 8.960 & 10.23 & 4.156 & 2.260 & 1.214 & 6.097 & 9.379 & 14.31 & 2.181 & 2.412 & 8.193 \\
0000001 & 33.33 & 7.292 & 2.173 & 1.199 & 9.753 & 36.97 & 12.42 & 6.156 & 3.056 & 9.303 & 77.74 & 42.07 & 6.190 & 4.516 & 21.86 \\
0000002 & 18.95 & 2.622 & 1.352 & 1.171 & 11.24 & 62.86 & 43.11 & 16.70 & 4.504 & 63.37 & 28.66 & 18.25 & 16.79 & 12.95 & 6.931 \\
0000003 & 16.05 & 6.228 & 1.487 & 1.159 & 4.794 & 20.46 & 13.18 & 9.044 & 5.225 & 27.59 & 22.70 & 21.03 & 3.799 & 3.967 & 33.19 \\
0000004 & 20.74 & 4.098 & 1.259 & 0.7186 & 25.40 & 43.40 & 53.23 & 7.534 & 4.215 & 77.68 & 61.60 & 28.62 & 2.476 & 1.399 & 43.55 \\
0000005 & 2.089 & 2.325 & 0.6713 & 0.6969 & 2.287 & 17.69 & 8.386 & 2.140 & 1.664 & 5.249 & 14.48 & 6.195 & 1.920 & 3.382 & 1.851 \\
0000006 & 15.38 & 9.743 & 1.219 & 1.312 & 13.27 & 31.87 & 3.710 & 4.300 & 2.357 & 3.268 & 30.64 & 22.79 & 3.376 & 4.493 & 8.719 \\
0000007 & 16.89 & 8.009 & 0.8079 & 1.010 & 6.053 & 13.98 & 6.446 & 2.422 & 2.008 & 5.632 & 14.88 & 10.71 & 2.393 & 4.754 & 5.271 \\
0000008 & 22.15 & 3.245 & 1.304 & 1.010 & 6.934 & 9.950 & 14.26 & 5.250 & 3.390 & 4.869 & 22.18 & 21.01 & 4.204 & 3.141 & 17.53 \\
0000009 & 5.119 & 3.754 & 0.7255 & 0.7261 & 3.260 & 33.39 & 12.20 & 5.575 & 2.616 & 11.29 & 13.29 & 12.39 & 4.915 & 3.132 & 1.785 \\
0000010 & 11.59 & 2.433 & 0.8929 & 0.9243 & 7.382 & 14.15 & 10.36 & 4.340 & 2.030 & 5.670 & 16.69 & 9.869 & 8.252 & 3.847 & 1.353 \\
0000011 & 1.734 & 2.948 & 0.6598 & 0.7596 & 1.716 & 9.671 & 8.274 & 5.233 & 1.829 & 8.059 & 6.788 & 12.46 & 5.067 & 4.178 & 2.499 \\
0000012 & 34.88 & 9.130 & 2.137 & 1.163 & 3.160 & 26.11 & 10.27 & 9.079 & 3.248 & 6.565 & 26.95 & 12.37 & 11.19 & 5.345 & 4.112 \\
\midrule
Mean & \textbf{15.79} & \textbf{5.126} & \textbf{1.193} & \textbf{0.9768} & \textbf{8.016} & \underline{25.44} & \underline{15.38} & 6.156 & \underline{2.874} & 18.05 & 26.62 & 17.85 & \underline{5.596} & 4.424 & \underline{12.07} \\
\bottomrule
\end{tabular}}
\end{table*}

\paragraph{Per-sequence results on YCBInEOAT.}
\label{supp:per-sequence-ycb}
Table~\ref{supp:tab:per-sequence-ycb} provides per-sequence results on YCBInEOAT. \method{} achieves the best mean ATE, ARE, RPE-t, and RPE-r despite using no depth or 3D model. 

\begin{table*}[t]
\centering
\caption{Per-sequence results on YCBInEOAT. All poses are relative to the first frame. \textbf{Bold}: best; \underline{underline}: second best.}
\label{supp:tab:per-sequence-ycb}
\footnotesize
\setlength{\tabcolsep}{3pt}
\resizebox{\textwidth}{!}{
\begin{tabular}{l|ccccc|ccccc|ccccc}
\toprule
& \multicolumn{5}{c|}{ProxyPose (one query)} & \multicolumn{5}{c|}{FoundationPose (track)} & \multicolumn{5}{c}{CoTracker 3 + DA3 (pixel level)} \\
\cmidrule(lr){2-6} \cmidrule(lr){7-11} \cmidrule(lr){12-16}
Sequence
& \metric{ATE}{mm} & \metric{ARE}{deg} & \metric{RPE-t}{mm} & \metric{RPE-r}{deg} & \metric{2D-dist}{px}
& \metric{ATE}{mm} & \metric{ARE}{deg} & \metric{RPE-t}{mm} & \metric{RPE-r}{deg} & \metric{2D-dist}{px}
& \metric{ATE}{mm} & \metric{ARE}{deg} & \metric{RPE-t}{mm} & \metric{RPE-r}{deg} & \metric{2D-dist}{px} \\
\midrule
00000\_obj12 & 20.63 & 5.962 & 2.695 & 1.455 & 1.742 & 25.33 & 10.17 & 4.730 & 1.921 & 8.497 & 28.39 & 13.65 & 4.111 & 3.977 & 1.796 \\
00001\_obj12 & 28.67 & 10.70 & 4.622 & 2.594 & 3.307 & 19.28 & 5.152 & 11.11 & 2.883 & 8.371 & 34.24 & 50.74 & 11.07 & 7.528 & 6.139 \\
00002\_obj2 & 16.73 & 5.804 & 7.495 & 2.557 & 13.93 & 38.92 & 11.16 & 19.88 & 5.500 & 19.58 & 58.99 & 27.96 & 17.26 & 16.34 & 17.26 \\
00003\_obj2 & 9.024 & 4.559 & 3.049 & 1.711 & 3.838 & 17.04 & 8.194 & 9.686 & 3.233 & 5.552 & 24.64 & 14.68 & 7.083 & 3.767 & 3.963 \\
00004\_obj5 & 30.02 & 9.716 & 2.219 & 1.456 & 18.83 & 32.68 & 7.413 & 16.82 & 2.961 & 8.589 & 42.00 & 14.48 & 19.58 & 6.311 & 8.232 \\
00005\_obj5 & 30.74 & 6.826 & 4.675 & 2.487 & 3.379 & 39.44 & 11.17 & 11.81 & 4.150 & 11.81 & 39.33 & 33.77 & 15.71 & 9.137 & 10.95 \\
00006\_obj3 & 104.4 & 79.94 & 9.789 & 7.858 & 26.94 & 216.6 & 7.264 & 18.14 & 4.840 & 12.48 & 182.7 & 42.32 & 13.36 & 6.334 & 9.999 \\
00007\_obj3 & 13.18 & 5.971 & 1.477 & 1.535 & 1.721 & 21.21 & 15.67 & 3.660 & 2.463 & 4.929 & 11.38 & 17.07 & 3.431 & 2.230 & 2.801 \\
00008\_obj4 & 17.27 & 6.155 & 2.947 & 2.019 & 2.772 & 41.20 & 90.98 & 4.731 & 4.263 & 18.95 & 7.130 & 6.372 & 4.377 & 2.901 & 1.887 \\
\midrule
Mean & \textbf{30.07} & \textbf{15.07} & \textbf{4.330} & \textbf{2.630} & \underline{8.496} & 50.19 & \underline{18.58} & 11.17 & \underline{3.579} & 10.97 & 47.65 & 24.56 & \underline{10.67} & 6.503 & \textbf{7.003} \\
\bottomrule
\end{tabular}}
\end{table*}

\paragraph{Quantitative results on synthetic dataset.}
\label{supp:synthetic-results}
Table~\ref{supp:tab:synthetic-results} presents results on our held-out synthetic benchmark. \method{} achieves the best ARE and RPE-r by a large margin, demonstrating accurate rotation tracking even in the absence of depth input. The synthetic scenes feature diverse objects, backgrounds, and motion trajectories with no overlap with the training set.

\begin{table*}[t]
\centering
\caption{Quantitative evaluation on the held-out synthetic dataset. All poses are relative to the first frame, with scale aligned at that frame. \textbf{Bold}: best; \underline{underline}: second best.}
\label{supp:tab:synthetic-results}
\footnotesize
\setlength{\tabcolsep}{2.4pt}
\begin{tabular}{l|ccc|ccccc}
\toprule
& \multicolumn{3}{c|}{Addl.\ Inputs} & \multicolumn{5}{c}{Synthetic Dataset} \\
\cmidrule(lr){2-4} \cmidrule(lr){5-9}
Method
& \vheader{3D Model}
& \vheader{Depth*}
& \vheader{Obj.\ Mask $\dagger$}
& \metric{ATE}{mm}
& \metric{ARE}{deg}
& \metric{RPE-t}{mm}
& \metric{RPE-r}{deg}
& \metric{2D-dist}{px} \\
\midrule
FoundationPose (track)
& \cb & \cb & & 892.2 & 28.87 & 99.06 & 5.470 & 43.82 \\
FoundationPose (registration)
& \cb & \cb & & 728.8 & 70.50 & 341.9 & 67.73 & 38.65 \\
Oryon
& & \cb & & 671.4 & 83.50 & 247.9 & 78.24 & 38.34 \\
\midrule
Any6D
& & \cb & \cb & 865.4 & 100.0 & 623.5 & 88.61 & 50.27 \\
One2Any
& & \cb & \cb & \textbf{261.4} & 59.10 & 102.8 & 20.41 & 40.08 \\
ConceptPose
& & \cb & \cb & \underline{269.9} & 105.2 & 323.7 & 115.8 & 50.25 \\
CoTracker 3 + DA3 (object level)
& & \cb & \cb & 787.4 & 38.67 & 120.2 & 10.10 & 26.87 \\
SpatialTracker V2 (object level)
& & \cb & \cb & 538.6 & 38.48 & 47.52 & 7.290 & 22.00 \\
BundleSDF
& & \cb & & 682.9 & 39.07 & 162.6 & 14.20 & 29.06 \\
\midrule
SpatialTracker V2 (pixel level)
& & \cb & & 453.2 & 40.90 & 50.65 & 19.58 & \textbf{12.56} \\
CoTracker 3 + DA3 (pixel level)
& & \cb & & 845.1 & 47.54 & 86.20 & 10.75 & 50.77 \\
\midrule
ProxyPose (one query)
& & & & 480.3 & \textbf{19.79} & \underline{29.92} & \textbf{1.920} & 15.17 \\
ProxyPose (two queries)
& & & \cb & 435.1 & \underline{21.53} & \textbf{29.61} & \underline{2.605} & \underline{15.03} \\
ProxyPose (three queries)
& & & \cb & 451.4 & 22.94 & 32.08 & 3.10 & 17.30 \\
\bottomrule
\end{tabular}
\vspace{2pt}

{\scriptsize *We use DA3 to provide depth inputs.}
{\scriptsize $\dagger$ Object masks required in one or more video frames.}

\end{table*}

\paragraph{Results with standard deviation.}
\label{supp:ho3d-ycbineoat-std}
Table~\ref{tab:std-quantitative-results} reproduces the main quantitative comparison from the main paper with per-sequence standard deviations included.

\begin{figure*}[t!]
\captionof{table}{Quantitative evaluation on HO3D and YCBInEOAT. All poses are relative to the first frame, with scale aligned at that frame. \textbf{Bold}: best; \underline{underline}: second best (lower is better for all metrics). Methods are grouped by input requirements, from most supervision (top) to least (bottom). \method{} (one query) is the only method that requires no 3D model, depth, or object mask.}
\label{tab:std-quantitative-results}
\footnotesize
\setlength{\tabcolsep}{2.4pt} 
\resizebox{\textwidth}{!}{
\begin{tabular}{l|ccccc|ccccc}
\toprule
& \multicolumn{10}{c}{Evaluation Benchmarks} \\
\cmidrule(lr){2-11}

Method
& \multicolumn{5}{c|}{HO3D}
& \multicolumn{5}{c}{YCBInEOAT} \\[-2pt]

\cmidrule(lr){2-6}
\cmidrule(lr){7-11}

& \metric{ATE}{mm}
& \metric{ARE}{deg}
& \metric{RPE-t}{mm}
& \metric{RPE-r}{deg}
& \metric{2D-dist}{px}

& \metric{ATE}{mm}
& \metric{ARE}{deg}
& \metric{RPE-t}{mm}
& \metric{RPE-r}{deg}
& \metric{2D-dist}{px}\\

\midrule

FoundationPose (registration)
& $53.2 \pm 35.4$ & $74.5 \pm 37.9$ & $37.1 \pm 18.8$ & $63.4 \pm 31.2$ & $37.1 \pm 20.2$
& $95.8 \pm 71.8$ & $118 \pm 56.6$ & $47.4 \pm 34.0$ & $71.1 \pm 31.0$ & $31.2 \pm 23.9$ \\

FoundationPose (track)
& $25.4 \pm 17.7$ & $15.4 \pm 14.4$ & $6.16 \pm 3.74$ & $2.87 \pm 1.23$ & $18.1 \pm 23.5$
& $50.2 \pm 61.7$ & $18.6 \pm 28.2$ & $11.2 \pm 5.79$ & $3.58 \pm 1.18$ & $11.0 \pm 5.15$ \\

Oryon
& $40.4 \pm 29.8$ & $38.0 \pm 22.6$ & $23.5 \pm 25.7$ & $23.7 \pm 25.3$ & $45.7 \pm 30.1$
& $41.2 \pm 56.8$ & $41.5 \pm 25.7$ & $20.9 \pm 15.6$ & $34.1 \pm 21.5$ & $25.6 \pm 20.0$ \\

\midrule

Any6D
& $64.6 \pm 56.6$ & $76.2 \pm 49.7$ & $41.1 \pm 40.3$ & $48.9 \pm 34.6$ & $56.2 \pm 56.5$
& $86.3 \pm 65.9$ & $67.5 \pm 35.8$ & $35.7 \pm 12.4$ & $37.4 \pm 10.0$ & $37.5 \pm 38.3$ \\

One2Any
& $52.0 \pm 34.3$ & $89.0 \pm 53.5$ & $12.9 \pm 9.70$ & $17.5 \pm 14.1$ & $55.3 \pm 43.3$
& $35.3 \pm 32.4$ & $34.0 \pm 32.5$ & $10.8 \pm 11.1$ & $10.3 \pm 9.49$ & $23.3 \pm 28.5$ \\

ConceptPose
& $34.2 \pm 28.8$ & $80.0 \pm 39.6$ & $38.9 \pm 31.2$ & $94.9 \pm 34.7$ & $38.1 \pm 27.4$
& $43.0 \pm 31.0$ & $68.4 \pm 41.5$ & $37.5 \pm 29.7$ & $75.7 \pm 40.4$ & $26.4 \pm 22.7$ \\

CoTracker 3 + DA3 (object level)
& $25.0 \pm 15.1$ & $19.6 \pm 10.3$ & $6.63 \pm 4.40 $ & $2.93 \pm 1.30 $ & $14.7 \pm 11.4$
& $44.8 \pm 56.8$ & $17.1 \pm 6.64$ & $10.8 \pm 5.71$ & $4.94 \pm 2.40$ & $7.84 \pm 6.91$ \\

SpatialTrackerV2 (object level)
& $25.9 \pm 13.8$ & $24.3 \pm 9.63$ & $1.87 \pm 0.811$ & $1.48 \pm 0.489$ & $25.1 \pm 17.6$
& $32.6 \pm 17.8$ & $22.4 \pm 19.0$ & $5.71 \pm 2.79$ & $3.23 \pm 1.69$ & $8.41 \pm 5.95$ \\

BundleSDF
& $23.2 \pm 18.5$ & $14.3 \pm 9.60 $ & $5.95 \pm 3.70 $ & $2.71 \pm 0.800 $ & $14.9 \pm 16.6$
& $42.1 \pm 59.3$ & $13.9 \pm 6.62$ & $10.3 \pm 5.23$ & $4.26 \pm 1.83$ & $\textbf{6.79} \pm 5.48$ \\

\midrule

SpatialTrackerV2 (pixel level)
& $26.1 \pm 20.2$ & $22.2 \pm 13.8$ & $1.90 \pm 0.838$ & $3.04 \pm 1.69$ & $25.9 \pm 28.7$
& $33.0 \pm 20.9$ & $23.3 \pm 8.27$ & $6.44 \pm 3.82$ & $8.50 \pm 9.18$ & $7.49 \pm 3.49$ \\

CoTracker 3 + DA3 (pixel level)
& $26.6 \pm 19.8$ & $17.9 \pm 9.80 $ & $5.60 \pm 4.40 $ & $4.42 \pm 3.20 $ & $\underline{12.1} \pm 14.0$
& $47.7 \pm 54.9$ & $24.6 \pm 14.9$ & $10.7 \pm 5.54$ & $6.50 \pm 2.08$ & $\underline{7.00} \pm 5.31$ \\

\midrule 

ProxyPose (one query)
& $\underline{15.8} \pm 10.5$ & $5.13 \pm 2.45$ & $\textbf{1.19} \pm 0.462$ & $0.977 \pm 0.225$ & $\textbf{8.02} \pm 6.59$
& $\underline{30.1} \pm 29.9$ & $15.1 \pm 24.0$ & $\textbf{4.33} \pm 2.79$ & $2.63 \pm 2.04$ & $8.50 \pm 9.47$ \\

ProxyPose (two queries)
& $\textbf{15.4} \pm 15.5$ & $\textbf{3.94} \pm 2.13$ & $\underline{1.22} \pm 0.616$ & $\underline{0.852} \pm 0.181$ & $12.3 \pm 18.5$
& $31.6 \pm 27.7$ & $\underline{7.76} \pm 4.78$ & $4.82 \pm 3.53$ & $\underline{2.27} \pm 0.922$ & $7.97 \pm 7.17$ \\

ProxyPose (three queries)
& $18.6 \pm 17.7$ & $\underline{4.30} \pm 2.48$ & $1.31 \pm 0.759$ & $\textbf{0.821} \pm 0.198$ & $14.6 \pm 20.4$ & $\textbf{26.3} \pm 18.9$ & $\textbf{6.48} \pm 2.26$ & $\underline{4.44} \pm 2.68$ & $\textbf{2.15} \pm 0.655$ & $7.77 \pm 7.09$ \\

\bottomrule
\end{tabular}
}
\vspace{2pt}
{\scriptsize *We use DA3 to provide depth inputs.}
{\scriptsize $\dagger$ Object masks required in one or more video frames.}
\vspace{-2em}
\end{figure*}

\subsection{Effect of Focal Length on Tracking Accuracy}
\label{supp:focal_length}

As noted in the main paper, \method{} assumes a known or coarsely estimated focal length $\focal$ for the PnP-based tracking stage.
To assess sensitivity to this assumption, we evaluate on HO3D (one query) with the ground-truth focal length scaled by factors of $1/2$ and $3/2$.
Table~\ref{supp:tab:focal_length} reports the results.

With the focal length halved, ARE increases from $5.1\degree$ to $15.2\degree$ and RPE-r from $0.98\degree$ to $1.88\degree$---both still competitive with most baselines in the main paper (Table~\ref{tab:quantitative-results}).
Translation and 2D-distance metrics degrade more substantially, as expected due to the coupling between focal length and depth.

\begin{table}[h]
\centering
\caption{Effect of focal-length error on tracking accuracy (HO3D, one query). The ground-truth focal length is $\focal$. Rotation metrics degrade gracefully under $\pm50\%$ focal-length error, while translation and 2D-distance metrics are more sensitive.}
\label{supp:tab:focal_length}
\footnotesize
\setlength{\tabcolsep}{4pt}
\begin{tabular}{l|ccccc}
\toprule
Focal length
 & \metric{ATE}{mm}
 & \metric{ARE}{deg}
 & \metric{RPE-t}{mm}
& \metric{RPE-r}{deg}
& \metric{2D-dist}{px} \\
\midrule
$1/2 \times f$  & 175.2 & 15.23 & 11.3 & 1.876 & 178.9 \\
$f$             & 15.79 & 5.126 & 1.193 & 0.9768 & 8.016 \\
$3/2\times f$   & 61.55 & 7.292 & 3.668 & 1.668 & 65.79\\
\bottomrule
\end{tabular}
\end{table}

\end{document}